%% file: main.tex
\documentclass[twocolumn,9pt,sigconf]{acmart}

\makeatletter
\def\@ACM@checkaffil{
    \if@ACM@instpresent\else
    \ClassWarningNoLine{\@classname}{No institution present for an affiliation}%
    \fi
    \if@ACM@citypresent\else
    \ClassWarningNoLine{\@classname}{No city present for an affiliation}%
    \fi
    \if@ACM@countrypresent\else
        \ClassWarningNoLine{\@classname}{No country present for an affiliation}%
    \fi
}
\makeatother

\usepackage{amsmath}
\usepackage{algorithmic}
\usepackage{graphicx}
\usepackage{textcomp}
\usepackage{xcolor}
\usepackage{multirow}
\usepackage{booktabs}
\usepackage{subcaption}
\usepackage{comment}
\usepackage{balance}
\usepackage{enumitem}
\usepackage{siunitx}
\newcommand{\blue}{\color{blue}}
\newcommand{\bluemark}[1]{{\color{blue}{#1}}}

\graphicspath{{figures/}}

\AtBeginDocument{%
  \providecommand\BibTeX{{%
    \normalfont B\kern-0.5em{\scshape i\kern-0.25em b}\kern-0.8em\TeX}}}

\copyrightyear{2023}
\acmYear{2023}
\setcopyright{acmlicensed}\acmConference[SenSys '23]{The 21st ACM Conference on Embedded Networked Sensor Systems}{November 12--17, 2023}{Istanbul, Turkiye}
\acmBooktitle{The 21st ACM Conference on Embedded Networked Sensor Systems (SenSys '23), November 12--17, 2023, Istanbul, Turkiye}
\acmPrice{15.00}
\acmDOI{10.1145/3625687.3625782}
\acmISBN{979-8-4007-0414-7/23/11}

\begin{document}

\title{MESEN: Exploit Multimodal Data to Design Unimodal Human Activity Recognition with Few Labels}
\author{Lilin Xu$^1$, Chaojie Gu$^{1, \ast}$\thanks{$^{\ast}$Chaojie Gu is the corresponding author.}, Rui Tan$^2$, Shibo He$^1$, Jiming Chen$^1$}
\affiliation{\institution{$^1$Zhejiang University, $^2$Nanyang Technological University}}
\email{Email: {lilinxu, gucj, s18he, cjm}@zju.edu.cn, tanrui@ntu.edu.sg}
\def \authors{Lilin Xu, Chaojie Gu, Rui Tan, Shibo He, Jiming Chen}

\renewcommand{\shortauthors}{Lilin Xu, Chaojie Gu, Rui Tan, Shibo He, Jiming Chen}

\begin{abstract}
Human activity recognition~(HAR) will be an essential function of various emerging applications.
However, HAR typically encounters challenges related to modality limitations and label scarcity, leading to an application gap between current solutions and real-world requirements.
In this work, we propose MESEN, a \underline{m}ultimodal-\underline{e}mpowered unimodal \underline{sen}sing framework, to utilize unlabeled multimodal data available during the HAR model design phase for unimodal HAR enhancement during the deployment phase.
From a study on the impact of supervised multimodal fusion on unimodal feature extraction, MESEN is designed to feature a multi-task mechanism during the multimodal-aided pre-training stage.
With the proposed mechanism integrating cross-modal feature contrastive learning and multimodal pseudo-classification aligning, MESEN exploits unlabeled multimodal data to extract effective unimodal features for each modality.
Subsequently, MESEN can adapt to downstream unimodal HAR with only a few labeled samples.
Extensive experiments on eight public multimodal datasets demonstrate that MESEN achieves significant performance improvements over state-of-the-art baselines in enhancing unimodal HAR by exploiting multimodal data.

\end{abstract}



\maketitle

\input{1_Introduction}
\input{2_RelatedWork}
\input{3_Preliminaries}

\input{5_Design}

\input{7_Evaluation}

\input{8_Discussion}

\input{9_Conclusion}
\input{10_acknowledge}

\bibliographystyle{ACM-Reference-Format}
\bibliography{ref}

\end{document}

%% file: 1_Introduction.tex

\section{Introduction}
\label{sec:intro}

In recent years, human activity recognition~(HAR) regains research attention due to the increasing applications and the potential performance leaps enabled by deep learning.
In particular, the recent advances in multimodal encoding in general artificial intelligence tasks, such as ImageBind~\cite{girdhar2023imagebind}, have implied that exploiting multimodal data for building HAR models is promising.
The prior studies~\cite{dawar2018convolutional,ma2019attnsense,lu2020milliego,ouyang2022cosmo} have shown the HAR performance improvements brought by multimodal data fusion.

\input{insert_figures/apllication_scenarios}

However, HAR in real-world scenarios still faces practical challenges and application gaps in exploiting multimodal data.
On the one hand, high annotation costs and label scarcity issues are widely present in practical HAR applications, resulting in scenarios where only a few labeled samples are available.
Manually annotating data is tedious and time-consuming. 
This issue becomes more severe in multimodal sensing, since annotating multimodal data requires correlating data across varied modalities and possessing knowledge of multiple modalities.
In comparison, unlabeled data are readily available and easier to access.
Potential solutions utilizing these easy-to-access unlabeled data can further boost data availability for HAR applications.
On the other hand, despite the growing prominence of multimodal research and deployment, unimodal HAR remains the most typical application paradigm.
In reality, many HAR applications are still deployed using a single modality. 
Even in scenarios where multiple sensors are available, there may still be requirements for unimodal HAR.
For instance, in smart home scenarios where mmWave radar sensors are deployed and smartwatches' built-in IMU sensors are available, users may not always be within the sensing range of the radar, resulting in situations of modality limitation. In these scenarios, although multimodal data can be collected, unimodal HAR remains accessible and important for users.
Thus, to achieve universal performance enhancement for HAR applications, it is important and meaningful to investigate the benefits of increasingly available multimodal data during the HAR model design phase on unimodal HAR during the deployment phase.

To this end, we aim to address a universal application situation as depicted in Figure \ref{fig:application_scenarios}, where the server has multimodal data for HAR model design while the user at the edge obtains a unimodal HAR model with few labeled samples. 
This scenario raises an essential question that remains to be studied: \textit{\textbf{how can we effectively exploit unlabeled multimodal data to improve the performance of unimodal HAR with few labels?}}

To answer the above question, we design MESEN, a multimodal-empowered unimodal sensing framework, to exploit multimodal data for designing unimodal HAR with few labels.
In this way, increasingly available unlabeled multimodal data can be exploited for effective unimodal feature extraction, thereby achieving universal enhancement for unimodal HAR.
From a study on the supervised multimodal fusion's effects on unimodal feature extraction, we observe that the correlations within temporally aligned multimodal samples and the distinct intra-modality spaces across different modalities are both vital for activity recognition.
Besides, we investigate the relationships between unimodal predicted probabilities and final fusion results.
In light of these observations, MESEN is designed to feature a multi-task mechanism during the multimodal-aided pre-training stage. 
By integrating cross-modal feature contrastive learning and multimodal pseudo-classification aligning, the mechanism exploits the correlations and relationships within unlabeled multimodal data, not only in the feature extraction stage's representation space but also in the pseudo-classification stage's representation space.
Equipped with effective unimodal features extracted during pre-training, MESEN then can adapt to downstream unimodal HAR with only a few labeled samples through fine-tuning.
This framework respects the single-modality constraint while effectively utilizing available unlabeled multimodal data.
 
We evaluate the performance of MESEN on eight multimodal datasets that encompass a range of modalities~(accelerometer, gyroscope, magnetometer, skeleton points, depth images, and mmWave radar), user scales, and human activities. 
Our evaluation indicates that MESEN achieves significant performance improvements on all datasets, yielding an average increase of 30.7\% accuracy and 34.5\% F1-score over supervised unimodal learning and at least an average increase of 25.2\% accuracy and 26.4\% F1-score over the contrastive learning baselines.

Our key contributions are summarized as follows:
\begin{itemize}[leftmargin=*]
\item By examining the performance of supervised multimodal fusion, we demonstrate the effects of utilizing multimodal data during training on unimodal feature extraction.
We further investigate the correlations and relationships within multimodal data in both the representation spaces of the feature extraction stage and the classification stage during multimodal training.

\item We propose to utilize the increasing availability of multimodal data to enhance unimodal HAR, given the widespread applicability of unimodal HAR in real-world scenarios.
MESEN\footnote{https://github.com/initxu/MESEN/}, a multimodal-empowered unimodal sensing framework, is designed to exploit unlabeled multimodal data for effective unimodal feature extraction by integrating cross-modal feature contrastive learning and multimodal pseudo-classification aligning.
With effective unimodal features, MESEN can adapt to downstream unimodal HAR with few labels.
\item We extensively evaluate the performance of MESEN on eight public multimodal datasets.
The results show that MESEN outperforms the state-of-the-art approaches in enhancing unimodal HAR performance by exploiting unlabeled multimodal data.
 
\end{itemize}

The rest of this paper is organized as follows. 
\S\ref{sec:related_work} reviews related studies. 
\S\ref{sec:preliminaries} presents the measurement study and motivation. 
\S\ref{sec:design} introduces the detailed design of MESEN. 
\S\ref{sec:evaluation} presents evaluation results. 
\S\ref{sec:discussion} discusses some related issues.
\S\ref{sec:conclusion} concludes this work.

%% file: insert_figures/apllication_scenarios.tex
\begin{figure}[t]
    \centering
    \includegraphics[width=0.48\textwidth]{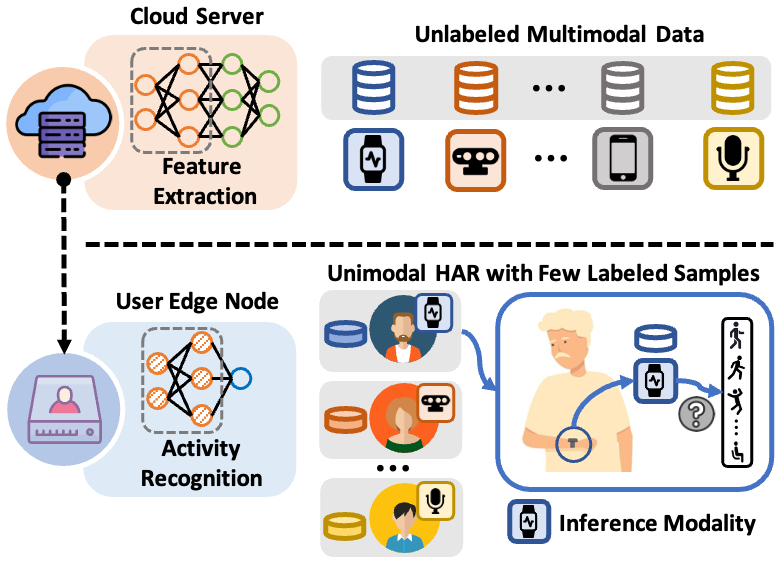}
    \caption{The application scenario of MESEN. Multimodal data are available on the server for HAR model design, while the user at the edge deploys unimodal HAR with few labels.}
    \label{fig:application_scenarios}
\end{figure}

%% file: 2_RelatedWork.tex
\vspace{-0.3em}
\section{Related work}
\label{sec:related_work}
In this section, we overview the research related to our work, demonstrating the research gaps that we aim to address. 

HAR is empowered by various modality sources, facilitating a wide range of applications including health monitoring~\cite{chen2013unobtrusive, dawar2018convolutional, he2022collaborative}, 
daily routine monitoring~\cite{gatica2019discovering, bi2017familylog}, and smart gym~\cite{radhakrishnan2020erica,rabbi2018virtual}.
However, real-world HAR often encounters the challenge of limited labeled data, primarily due to the high costs and labor intensity associated with the data labeling process, especially for the modalities beyond RGB videos which are in general non-interpretable by people. 
Consequently, the study of HAR with limited labeled data has garnered significant research interest in recent years.

Generative methods~\cite{yao2018sensegan,chen2023hmgan} craft data or labels from existing data to mitigate the issue of label scarcity.
SenseGAN~\cite{yao2018sensegan} utilizes limited labeled supervision and abundant unlabeled data. It features a generator producing sensing data with random labels, a classifier producing labels for unlabeled data, and a discriminator discriminating real labeled samples and partially generated samples.
HMGAN~\cite{chen2023hmgan}, as a data augmentation technique, employs multiple generators to generate multimodal data from limited labeled data to enlarge the training set.

Besides, different from generative methods, there are studies on employing self-supervised learning techniques to exploit available unlabeled data directly to their advantage.
Depending on the modalities involved in the model design and deployment phases, existing studies can be divided into the categories of single-modality~\cite{saeed2019multi,sheng2022facilitating} and multi-modality~\cite{ma2021unsupervised,haresamudram2021contrastive,xu2021limu,deldari2022cocoa,ouyang2022cosmo}.

IMU-based methods~\cite{saeed2019multi,ma2021unsupervised,haresamudram2021contrastive,xu2021limu} have been extensively studied in recent years. 
While TPN~\cite{saeed2019multi} is designed to deal with unimodal IMU data by recognizing eight different data transformations applied to the raw accelerometer signal, the approaches in~\cite{ma2021unsupervised,haresamudram2021contrastive,xu2021limu} are capable of processing data from multiple IMU sensors by performing the data-level fusion among modalities with a relatively small gap.
Multi-task deep cluster~\cite{ma2021unsupervised} employs a framework with three tasks trained iteratively to obtain effective representation.
CPCHAR~\cite{haresamudram2021contrastive} utilizes the Contrastive Predictive Coding (CPC) framework to capture the temporal structure of the accelerometer and gyroscope data.
LIMUBert~\cite{xu2021limu} proposes an adaptive BERT-like self-supervised task to extract generalizable features from unlabeled IMU sensor data.
RadarAE~\cite{sheng2022facilitating} adapts the idea of Masked Autoencoders~(MAE) to radar sensing with unlabeled radar data.
These approaches are specifically designed for a single modality or modalities with similar natures, which prevents them from being easily ported to other modalities or handling heterogeneous multimodal data.
We propose a universal framework that is capable of handling various modalities without any extra changes.

With the development of contrastive learning and its excellent representation learning performance in multiple fields, such as computer vision~\cite{chen2020simple,tian2020contrastive} and natural language~\cite{fang2020cert,wang2021cline}, this effective idea is introduced to HAR to deal with unlabeled multimodal data.
COCOA~\cite{deldari2022cocoa} performs contrastive learning between features extracted from multisensor data for fusion.
Cosmo~\cite{ouyang2022cosmo} designs a multimodal feature fusion contrastive method to fully use multimodal synergies within heterogeneous multimodal data.
These solutions focus on improving the performance of multimodal HAR for scenarios where multimodal data are available.
Thus, they cannot be directly applied to our target unimodal HAR application scenarios where only a single modality is available during the deployment phase.
\input{insert_figures/framework_compare}

Different from the above single-modality or multi-modality methods, our method, MESEN, aims to utilize increasingly available unlabeled multimodal data to achieve universal enhancement for unimodal HAR.
Thus, the proposed framework operates in a multi-to-unimodal mode as shown in Figure \ref{fig:framework_compare} (c).
Although similar two-stage training frameworks~(Figure \ref{fig:framework_compare} (a) \& (b)) are used in previous works~\cite{haresamudram2021contrastive,sheng2022facilitating,ouyang2022cosmo} to address label scarcity issues for unimodal or multimodal HAR, MESEN significantly differs from them due to the specific application scenarios involving modality limitations during the HAR model design and the deployment phases.

%% file: insert_figures/framework_compare.tex
\begin{figure}[t]
    \centering
    \hspace{0.00\textwidth}
    \includegraphics[width=0.48\textwidth]{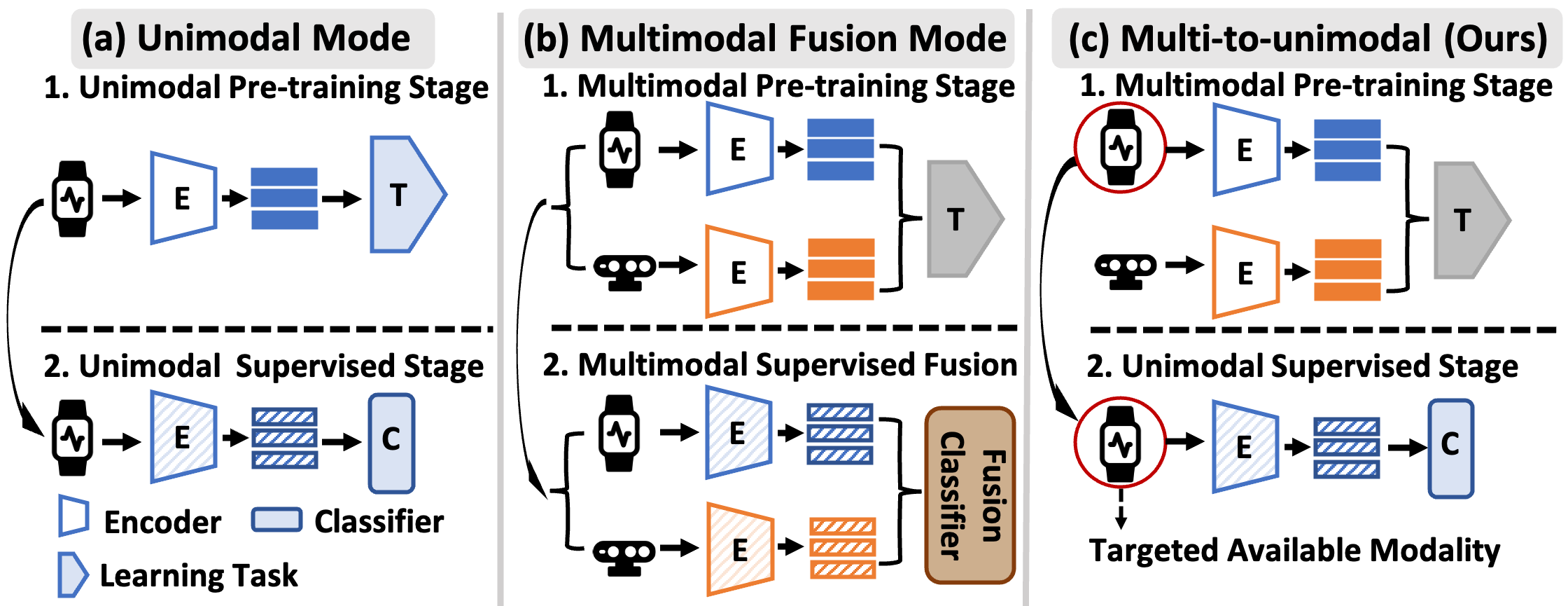}
    \caption{\textbf{{(a) \& (b):}} Prior works~\cite{haresamudram2021contrastive,sheng2022facilitating,ouyang2022cosmo} designed with label scarcity include the unimodal mode and the multimodal fusion mode.
    \textbf{{(c):}} MESEN operates in a multi-to-unimodal mode to improve unimodal HAR performance by exploiting unlabeled multimodal data.}
    \label{fig:framework_compare}
\end{figure}

%% file: 3_Preliminaries.tex
\section{Motivation}
\label{sec:preliminaries}
In this section, we conduct a measurement study to understand modality differences and investigate the impact of multimodal fusion on unimodal feature extraction.
The observations motivate the design of MESEN, a framework that addresses practical multi-to-unimodal HAR application scenarios with few labels.
\input{insert_figures/uci_cm.tex}

\subsection{Measurement Study}
\label{sec:multimodal_assistance}
Different modalities capture different aspects of a process. 
For instance, accelerometer measures linear acceleration; gyroscope measures angular velocity and is thus sensitive to changes in orientation. 
As a result, they are sensitive to different types of human activities.
We conduct experiments on the UCI dataset~\cite{reyes2016transition} comprising accelerometer and gyroscope data for six activities, with four users' data for training and the rest for validation and testing. 
We use 1D-CNN~\cite{tang2020rethinking} as modality encoders and a single linear layer as classifier heads.
Figure \ref{fig:uci_cm} shows that when employed individually, each modality yields better performance on certain types of activities. 
Gyroscope demonstrates better recognition performance for dynamic activities compared with stationary ones, due to its sensitivity to orientation changes and angular velocity.
Accelerometer outperforms gyroscope in recognizing stationary activities, due to its ability to measure static forces like gravity.

However, the gyroscope's relatively worse performance in identifying stationary activities does not mean it lacks relevant information for distinguishing these activities.
Instead, the relevant valuable information and features are present but not effectively captured during the training. 
Indeed, the results show that gyroscope data can still recognize a part of these activities effectively.

Actually, the \textbf{\textit{free yet accessible}} features, which are effective for recognition, can be unearthed when another modality is present during training, as in multimodal fusion.
Multimodal fusion has been considered in HAR~\cite{yao2017deepsense,ma2019attnsense,arigbabu2020entropy,jain2017human} and has shown notable performance improvements when the multimodal features or the predicted probabilities are properly fused.
We investigate the impact of the assistance provided by an additional modality during training by studying unimodal performance under supervised multimodal fusion training.
We utilize a score-level fusion method~\cite{jain2017human}, which applies a weighted fusion on unimodal predicted probabilities obtained from each modality.
On the one hand, as Figure \ref{fig:uci_cm} shows, multimodal fusion improves the recognition performance.
On the other hand, the fusion also aids in extracting effective unimodal features that might otherwise remain undetected during unimodal training.
With t-SNE visualization~\cite{van2008visualizing}, we analyze the gyroscope features extracted by the modality encoder under three conditions: without training, post unimodal training, and during multimodal fusion.
Figure \ref{fig:gyr_features} shows that when accelerometer data are present during training, the gyroscope features form more distinct clusters compared with other conditions.
This suggests that effective unimodal features can be better extracted with the aid of another modality under supervised multimodal fusion.

\input{insert_figures/gyr_features}
\input{insert_figures/cosmo_features}
\subsection{Problem Statement}

Based on the experiments, \S\ref{sec:multimodal_assistance} gives insights into the effectiveness of unimodal feature extraction achieved during supervised multimodal fusion training.
With these insights, we aim to apply such multimodal assistance for unimodal HAR with unlabeled multimodal data.

The scenarios we focus on have two major differences from supervised multimodal HAR, i.e., exploiting unlabeled multimodal data and enhancing unimodal HAR.
Firstly, real-world scenarios often face the challenge of lacking annotations, but unlabeled temporally aligned multimodal data are easily accessible.
By utilizing unlabeled multimodal data, the availability of data can be significantly increased, thus avoiding the constraints imposed by reliance on labeled data.
Secondly, driven by the ubiquity of unimodal HAR application scenarios, we focus on utilizing unimodal data for activity recognition during the deployment phase rather than applying multimodal HAR. Our aim is to improve the performance of unimodal HAR with available multimodal data.

The most related work is Cosmo~\cite{ouyang2022cosmo}, which is capable of handling unlabeled multimodal data for HAR.
However, it cannot be directly applied to our target application scenarios as it is designed to fuse multimodal data for scenarios where multimodal data are available during the deployment phase of HAR applications.
Figure \ref{fig:cosmo_features} shows that the unimodal features extracted during Cosmo's pre-training stage do not exhibit the same clear clustering characteristic as the fusion features.
This indicates that Cosmo is designed to utilize unlabeled multimodal data for extracting features that are beneficial to subsequent multimodal fusion.
In contrast, our target scenarios require directly extracting effective unimodal features for downstream unimodal HAR.
Consequently, we design MESEN to exploit unlabeled multimodal data and achieve effective unimodal feature extraction with multimodal assistance, thereby enhancing unimodal HAR with few labels.

%% file: insert_figures/uci_cm.tex
\begin{figure}[t]
    \centering
    \includegraphics[width=0.47\textwidth]{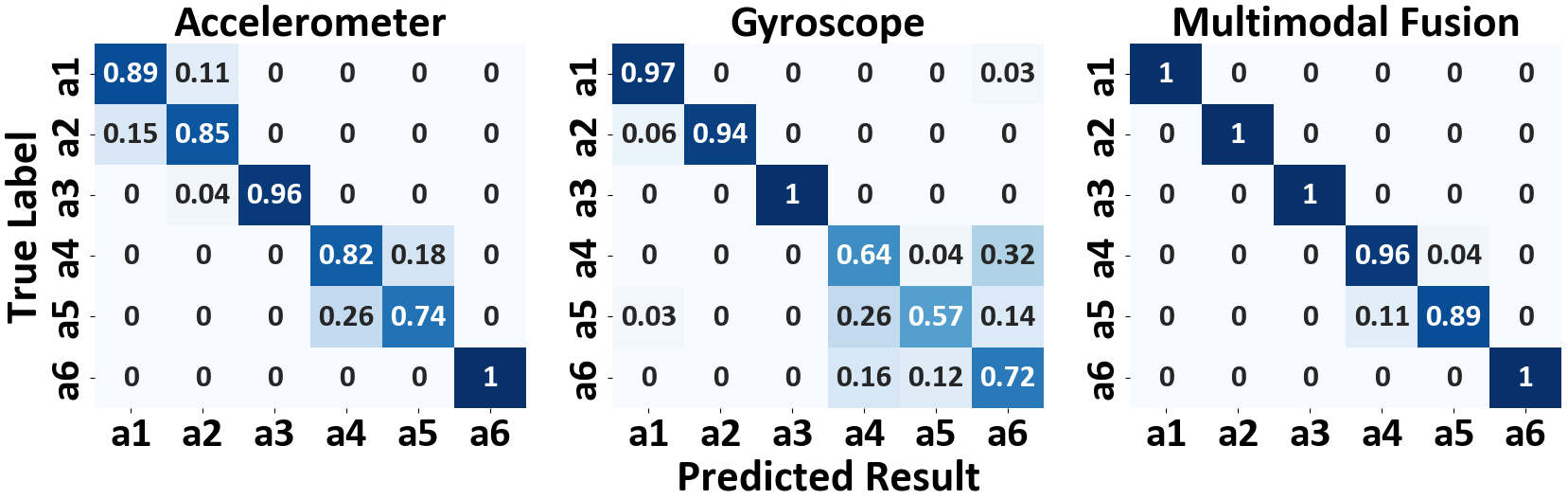}
    \caption[Caption for LOF]%
    {Unimodal and multimodal recognition results on the UCI dataset. 
    Activities from $a1$ to $a3$ are walking-related while the rest are stationary activities.
    }
    \label{fig:uci_cm}
\end{figure}

%% file: insert_figures/gyr_features.tex
\begin{figure}[t]
    \centering
    \captionsetup{skip=8pt}
    \includegraphics[width=0.48\textwidth]{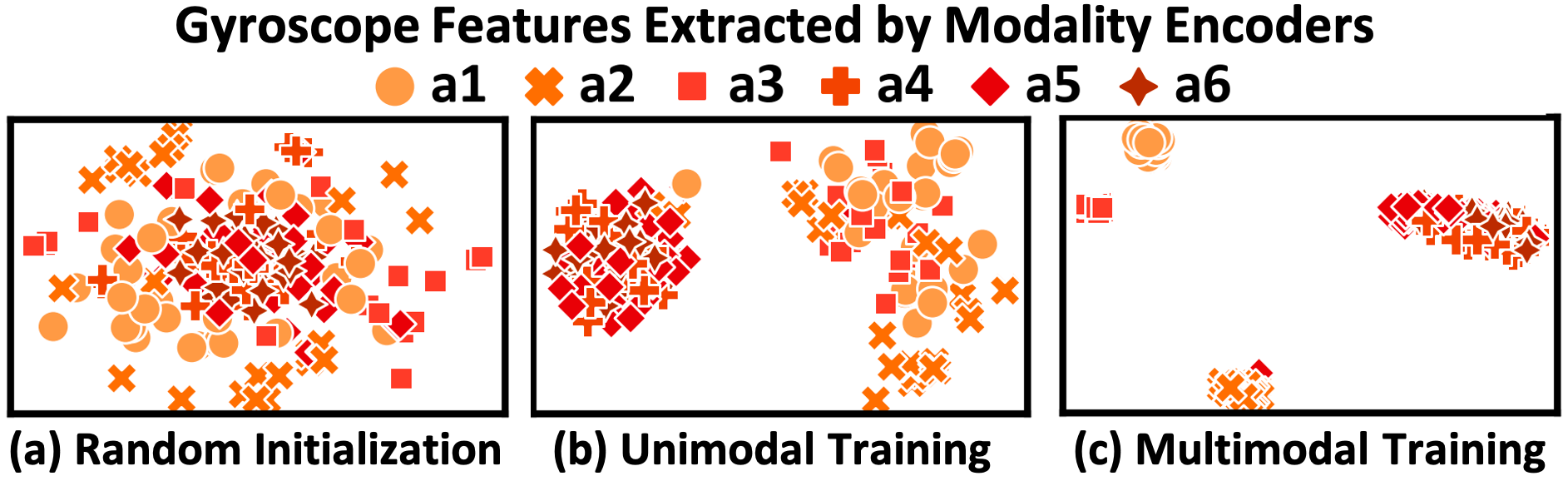}
    \caption{The visualization of extracted gyroscope features under three conditions.}
    \label{fig:gyr_features}
\end{figure}

%% file: insert_figures/cosmo_features.tex
\begin{figure}[t]
    \centering
    \captionsetup{skip=8pt}
    \includegraphics[width=0.48\textwidth]{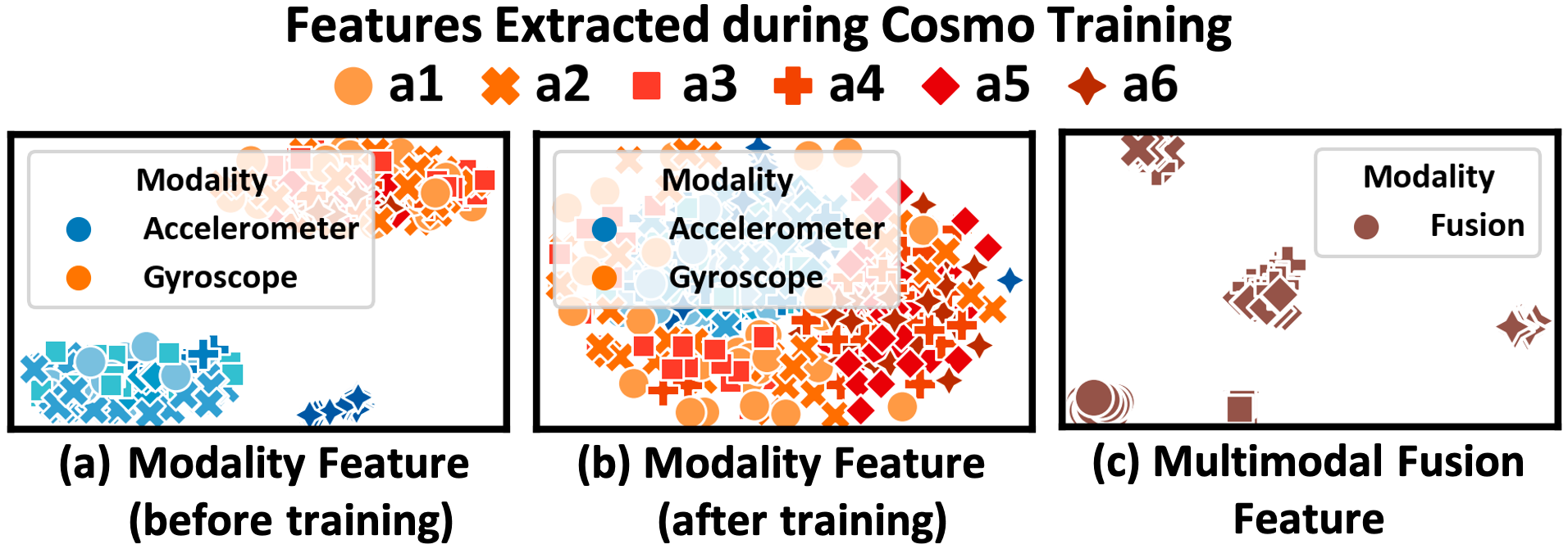}
    \caption{The unimodal features extracted by Cosmo are beneficial to subsequent multimodal fusion instead of unimodal recognition.}
    \label{fig:cosmo_features}
\end{figure}

%% file: 5_Design.tex
\section{MESEN Design}
\label{sec:design}
In this section, we indicate the specific application scenarios that our work focuses on. 
Subsequently, we demonstrate the design of MESEN in detail.

\subsection{Overview}
\noindent $\blacksquare$ \textbf{Problem formulation:} 
The unlabeled multimodal data with $N$ modalities available are defined as $\mathcal{D}_{m}=\{\mathcal{D}^{M_1}, \ldots, \mathcal{D}^{M_N}\}$, where $\mathcal{D}^{M_i}=\{x^{M_i}_1, \ldots, x^{M_i}_{n_{M_i}}\}$ and $n_{M_i}$ denotes the sample number.
$x^{M_i}_k$ and $x^{M_j}_k$ denote temporally aligned paired samples recording the same activity process.
The unimodal data during the deployment phase are defined as $\mathcal{D}_{u}=\{\mathcal{D}^{M_{a}}\}$, where $\mathcal{D}^{M_a}=\{\hat{x}^{M_a}_1, \ldots, \hat{x}^{M_a}_{m_{M_a}}, x^{M_a}_1, \ldots, x^{M_a}_{n_{M_a}}\}$.
It involves labeled data $\hat{x}^{M_a}$ with a small value of $m_{M_a}$ of the available modality $M_{a} \in \{M_1,\ldots, M_N\}$.
MESEN is designed to exploit multimodal data from $\mathcal{D}{m}$ to enhance the unimodal HAR performance on $\mathcal{D}{u}$. 

\noindent $\blacksquare$ \textbf{Framework overview:} Figure \ref{fig:design} demonstrates the oveview design of MESEN.
For clarity, we use two modalities $M_a$ and $M_b$ as the illustration in the figure and the subsequent design description.
MESEN comprises the multimodal-aided pre-training stage and the unimodal fine-tuning stage.
During pre-training, MESEN aims to train modality encoders for effective unimodal feature extraction, utilizing unlabeled multimodal data and modality relationships.
During fine-tuning, the encoder of the available modality is fine-tuned with a few labeled samples and then used for unimodal HAR.

\input{insert_figures/framework}

\subsection{Multimodal-aided Pre-training Stage}
As mentioned in \S\ref{sec:preliminaries}, supervised multimodal fusion can aid unimodal feature extraction.
To achieve this assistance with unlabeled multimodal data, we further investigate the correlations and relationships within multimodal data in both the representation spaces of the feature extraction stage and the classification stage.
Based on the insights observed from supervised multimodal fusion, we develop a multi-task mechanism, integrating cross-modal feature contrastive learning and multimodal pseudo-classification aligning. With the mechanism, unimodal features can be effectively extracted from unlabeled multimodal data for subsequent unimodal HAR.

\input{insert_figures/modality_consistent}
\subsubsection{Cross-modal Feature Contrastive Learning.}
Paired multimodal data are records of the same process, suggesting their inherent correlations, while they capture different aspects of the process due to their distinct physical properties.
Thus, we investigate the representation space of the feature extraction stage for both the correlations and differences between modalities.
We observe there are inter-modality correlations in paired multimodal data~(as depicted in Figure \ref{fig:modality_consistent}) and distinct intra-modality spaces reflecting the differences between modalities~(as depicted in Figure \ref{fig:feature_space}).
Based on the observations, we employ a cross-modal feature contrastive learning method rather than applying contrastive learning directly.
On the one hand, this method emphasizes the similarity between paired multimodal features to capture inter-modality correlations. 
On the other hand, it maintains the modality differences by excluding consideration of dissimilarity within intra-modality when maximizing the dissimilarity between non-paired multimodal features.

\noindent $\blacksquare$ \textbf{Inter-modality correlation:} 
Multiple modalities can acquire crucial correlated information when recording the same activity process. 
For instance, the left portion of Figure \ref{fig:modality_consistent} shows temporally aligned paired samples of two modalities~(accelerometer and gyroscope), which record the same \textit{Walking} activity process.
The paired samples exhibit similar variation patterns due to the rhythmic nature of walking.
These correlated patterns play a vital role in recognition. 
To highlight significant regions in the input sensor data that contribute to the final prediction in multimodal fusion, we apply Gradient-weighted Class Activation Mapping (Grad-CAM)~\cite{selvaraju2017grad} to the last convolutional layer of each modality encoder individually after supervised multimodal fusion training.
As shown in the right portion of Figure \ref{fig:modality_consistent}, there is a correlation between the important regions across modalities.
Actually, such correlated information is ubiquitously present in paired multimodal data. 
On the UCI dataset, we conduct canonical correlation analysis on data from the same participant.
The results show an average correlation of 0.600 between paired multimodal samples and an average correlation of 0.313 between non-paired samples, indicating that paired samples hold more correlated information.

We aim to utilize the correlated information in multimodal data to enhance unimodal feature extraction.
An effective strategy to capture the correlations involves maximizing the similarity of paired multimodal features while maximizing the dissimilarity of non-paired features.

\noindent $\blacksquare$ \textbf{Intra-modality space:} 
Different modalities have distinct physical properties.
While non-paired features in multimodal scenarios encompass both inter-modality and intra-modality types, it is inappropriate to maximize the dissimilarity of both types in our task due to the modality differences.
Figure \ref{fig:feature_space} (a) demonstrates distinct intra-modality spaces reflecting the physical differences between modalities.
Furthermore, Figure \ref{fig:feature_space} (b) shows that at the level of latent feature representation in supervised multimodal fusion, features from the same modality tend to cluster together, forming distinct modality-specific spaces for different modalities.
This indicates that the intra-modality spaces related to modality properties are crucial for activity recognition.
Therefore, it is beneficial to maintain distinct intra-modality spaces during pre-training.
However, as shown in Figure \ref{fig:feature_space} (c), maximizing the dissimilarity of both inter-modality and intra-modality non-paired features during contrastive learning can lead to distortion of the distinct intra-modality feature spaces.
This is due to the neglect of the distinct properties of different modalities and treating them equally in contrastive learning.

To prevent this kind of distortion and maintain modality differences, we only maximize the dissimilarity of inter-modality non-paired features without maximizing the dissimilarity between intra-modality features.

\input{insert_figures/feature_space}
\noindent $\blacksquare$ \textbf{Design for correlation capturing \& difference maintaining:} 
Given the observations above, we design cross-modal feature contrastive learning to capture inter-modality correlations and maintain distinct intra-modality spaces during training.

As shown in Figure \ref{fig:design}, $\mathbf{x^a}$ and $\mathbf{x^b}$ denote temporally aligned paired samples from modalities $M_a$ and $M_b$. These samples are processed through respective modality encoders~$\boldsymbol{f}(\cdot)$ and modality projectors~$\boldsymbol{g}(\cdot)$, yielding features $\mathbf{z^a} \in \mathbb{R}^{N_{fc}}$ and $\mathbf{z^b} \in \mathbb{R}^{N_{fc}}$, which can be expressed by 
\begin{equation}
\begin{split}
\mathbf{h^a}=\boldsymbol{f}^{M_a}(\mathbf{x^a}),
\mathbf{\hat{h}^a}={\boldsymbol{g}^{M_a}(\mathbf{h^a})},
\mathbf{z^a}=\boldsymbol{M_{Norm}}(\mathbf{\hat{h}^a});\\
\mathbf{h^b}=\boldsymbol{f}^{M_b}(\mathbf{x^b}),
\mathbf{\hat{h}^b}={\boldsymbol{g}^{M_b}(\mathbf{h^b})},
\mathbf{z^b}=\boldsymbol{M_{Norm}}(\mathbf{\hat{h}^b}),
\label{eq:feature}
\end{split}
\end{equation}
where the modality features $\mathbf{\hat{h}}$ are mapped and normalized within the batch by $\boldsymbol{M_{Norm}}(\cdot)$ to $\mathbf{z}$ for contrastive learning.
During multimodal pre-training, every input pairs $(\mathbf{x^a}_i,\mathbf{x^b}_i)$ produces a pair of feature vectors $(\mathbf{z^a}_i,\mathbf{z^b}_i)$.
Across $N$ input pairs in the mini-batch $\mathcal{B}$, we get $\mathcal{B}_{z}^{a}=\{\mathbf{z^a}_1, \ldots, \mathbf{z^a}_N\}$ and $\mathcal{B}_{z}^{b}=\{\mathbf{z^b}_1, \ldots, \mathbf{z^b}_N\}$ for the computation of the contrastive loss.

\input{insert_figures/modalitycontrastive}
Unlike the single-modality contrastive loss~\cite{chen2020simple}, which constructs one positive pair and $2N-2$ negative pairs for each sample within a mini-batch, we obtain one inter-modality positive pair, $N-1$ inter-modality negative pairs and $N-1$ intra-modality negative pairs.
Specifically, for the sample $\mathbf{x^a}_i$, as shown in Figure \ref{fig:cmcl}, we compute the contrastive loss based on the positive set $\mathbf{P^a_i}=\{\mathbf{z^b}_i\}$ and the inter-modality negative set $\mathbf{N^{({a}\rightarrow{b})}_i}=\{\mathbf{z^b}_j | \mathbf{z^b}_j \in \mathcal{B}_{z}^{b}, j \neq i\}$, while ignoring the intra-modality negative set $\mathbf{N^{({a}\rightarrow{a})}_i}=\{\mathbf{z^a}_j | \mathbf{z^a}_j \in \mathcal{B}_{z}^{a}, j \neq i\}$. Consequently, the $M_a$-to-$M_b$ contrastive loss $L^{({a}\rightarrow{b})}_{i}$ for the sample $\mathbf{x^a}_i$ can be expressed by
\begin{equation}
\resizebox{0.9\hsize}{!}{
$
L^{({a}\rightarrow{b})}_{i} = -\log\frac{\exp(\mathbf{z^a}_i \cdot \mathbf{z^b}_i/\tau)}{\exp(\mathbf{z^a}_i \cdot \mathbf{z^b}_i/\tau) + \sum_{\mathbf{z^b}_j \in \mathbf{N^{({a}\rightarrow{b})}_i}}\exp(\mathbf{z^a}_i \cdot \mathbf{z^b}_j/\tau)},
$
}
\label{eq:a2b_cl}
\end{equation}
where $\tau \in \mathbb{R}^+$ denotes the temperature parameter.
Indeed, Eq. \ref{eq:a2b_cl} can be interpreted as the loss of seeking to correctly identify $(\mathbf{z^a}_i, \mathbf{z^b}_i)$ as the temporally aligned pair, while treating the rest as negatives.
Additionally, we can obtain the $M_b$-to-$M_a$ contrastive loss $L^{({b}\rightarrow{a})}_{i}$ for the sample $\mathbf{x^b}_i$ in the same way, it can be expressed by
\begin{equation}
\resizebox{0.9\hsize}{!}{
$
L^{({b}\rightarrow{a})}_{i} = -\log\frac{\exp(\mathbf{z^b}_i \cdot \mathbf{z^a}_i/\tau)}{\exp(\mathbf{z^b}_i \cdot \mathbf{z^a}_i/\tau) + \sum_{\mathbf{z^a}_j \in \mathbf{N^{({b}\rightarrow{a})}_i}}\exp(\mathbf{z^b}_i \cdot \mathbf{z^a}_j/\tau)}.
$
}
\label{eq:b2a_cl}
\end{equation}

The cross-modal feature contrastive loss $L_{\text{CMF}}$ for the mini-batch $\mathcal{B}$ of $N$ paired samples can be expressed as
\begin{equation}
L_{\text{CMF}}=\frac{1}{N}\sum^{N}_{i=1}(\alpha L^{({a}\rightarrow{b})}_{i} + \beta L^{({b}\rightarrow{a})}_{i} ),
\label{eq:cmf_cl}
\end{equation}
where $\alpha > 0$ and $\beta > 0$ are weights measuring the importance of different modalities contributing to $L_{\text{CMF}}$.
To adapt to various modality combinations, we assign equal weights to different modalities by setting $\alpha = \beta = 0.5$.

\subsubsection{Multimodal pseudo-classification aligning.}
\label{sec:multimodal_pca}
After utilizing the multimodal correlations in the representation space of the feature extraction stage, there is still a lack of a pre-training task that is directly related to the downstream recognition task in the design.
Besides, the multimodal correlations existing in the representation space of the classification stage remain unexploited.
Thus, we employ multimodal pseudo-classification aligning.
On the one hand, the pseudo-classification task can be a prompt for the downstream recognition task.
On the other hand, it can utilize relationships within multimodal predicted probabilities in the classification stage's representation space.
\input{insert_figures/decision_cls_space}

\noindent $\blacksquare$ \textbf{Task prompt:} 
Introducing a pseudo-classification task into the pre-training stage provides a beneficial prompt for downstream recognition.
Given that activity recognition is essentially a classification task, employing a pseudo-classification task during pre-training can benefit feature extraction, and enhance the effectiveness of the fine-tuning process even if there are only a few labeled samples available.

\noindent $\blacksquare$ \textbf{Multimodal predicted probabilities alignment:} 
The alignment measures the degree of similarity between the unimodal prediction probabilities of paired samples.
As discussed in \S\ref{sec:preliminaries}, the fusion of unimodal predicted probabilities can guide unimodal feature extraction. 
To utilize pseudo-classification results without the fusion step, we investigate the relationships between the multimodal predicted probabilities alignment and the final fusion prediction results under supervised multimodal fusion.

During supervised multimodal fusion, unimodal predicted probabilities~($\mathbf{{y}^a}$ and $\mathbf{{y}^b}$) are individually mapped from modality features through the classifier head. The fusion result $\mathbf{{y}}$ is obtained by combining $\mathbf{{y}^a}$ and $\mathbf{{y}^b}$.
Our observations show that for $\mathbf{{y}}$ that is correctly classified (as depicted at the middle top portion in Figure \ref{fig:decision_cls_space}), $\mathbf{{y}^a}$ and $\mathbf{{y}^b}$ exhibit a high degree of alignment.
Conversely, for the misclassified fusion result, $\mathbf{{y}^a}$ and $\mathbf{{y}^b}$ vary significantly from each other~(as depicted at the middle bottom portion in Figure \ref{fig:decision_cls_space}).
We measure the Euler distance between $\mathbf{{y}^a}$ and $\mathbf{{y}^b}$ on the UCI dataset, revealing that the average distance associated with misclassified fusion results is 0.842, which is 2.38 times the average distance of 0.354 related to correctly classified results.
If the unimodal predicted probabilities of paired samples are significantly different, it suggests that at least one of the results provides a less reliable classification prediction, which adversely affects the final fusion result.

Based on the above observations, as shown in Figure \ref{fig:pccl}, our objective is to ensure that the pseudo-predicted probabilities~($\mathbf{\hat{y}_{i}^a}$, $\mathbf{\hat{y}_{i}^b}$) obtained from paired multimodal samples~($\mathbf{x_{i}^a}$, $\mathbf{x_{i}^b}$) are classified into the same pseudo-class, while those derived from non-paired samples should be classified into different pseudo-classes.

\input{insert_figures/classificationcontrastive}
\noindent $\blacksquare$ \textbf{Design for pseudo-classification aligning:} 
Given the insights above, we employ multimodal pseudo-classification aligning. It first implements pseudo-classification on modality features and then utilizes the relationships within multimodal pseudo-class probability probabilities.

Firstly, we apply a $N_{cls}$-way classification by using the pseudo-classification head $\boldsymbol{\hat{c}}(\cdot)$, where $N_{cls}$ is the number of activity categories which can be easily obtained with no extra effort.
The pseudo-predicted probabilities $\mathbf{\hat{y}^a}=\boldsymbol{\hat{c}}(\mathbf{\hat{h}^a})$ and $\mathbf{\hat{y}^b}=\boldsymbol{\hat{c}}(\mathbf{\hat{h}^b})$ are obtained from the modality features $\mathbf{\hat{h}^a}$ and $\mathbf{\hat{h}^b}$ individually. 
Then, we directly utilize $\mathbf{\hat{y}^a}$ and $\mathbf{\hat{y}^b}$ instead of combining them as in supervised multimodal fusion.

Specifically, within the mini-batch $\mathcal{B}$, we obtain results of pseudo-classification~$\mathbf{\hat{Y}^a}$ and $\mathbf{\hat{Y}^b}$, both of which belong to the space $\mathbb{R}^{N \times N_{cls}}$ and can be expressed as
\begin{equation}
\begin{split}
\mathbf{\hat{Y}^{a}}=\begin{bmatrix} \mathbf{\hat{y}^{a}_1} \\ \ldots \\ \mathbf{\hat{y}^{a}_N} \end{bmatrix} = \begin{bmatrix} \mathbf{\hat{q}^{a}_{1}} \ldots \mathbf{\hat{q}^{a}_{N_{cls}}} \end{bmatrix},
\mathbf{\hat{Y}^{b}}=\begin{bmatrix} \mathbf{\hat{y}^{b}_1} \\ \ldots \\ \mathbf{\hat{y}^{b}_N} \end{bmatrix} = \begin{bmatrix} \mathbf{\hat{q}^{b}_{1}} \ldots \mathbf{\hat{q}^{b}_{N_{cls}}} \end{bmatrix}.
\label{eq:pc_y}
\end{split}
\end{equation}
The $i$-th element in the vector $\mathbf{\hat{y}}$ represents the probability that $\mathbf{{x}}$ is the $i$-th pseudo-class. The $i$-th column $\mathbf{\hat{q}_{i}}$ in the matrix $\mathbf{\hat{Y}}$ represents the $i$-th pseudo-class, mapping that which samples in $\mathcal{B}$ are classified into the $i$-th pseudo-class.
Ideally, paired multimodal samples $\mathbf{x^a}$ and $\mathbf{x^b}$ should be classified into the same pseudo-class.
As a result, $\mathbf{\hat{q}^{a}_{i}}$ and $\mathbf{\hat{q}^{b}_{i}}$, both denoting the $i$-th pseudo-class results, should be as similar as possible.
In contrast, the dissimilarity between the $i$-th pseudo-class $\mathbf{\hat{q}_{i}}$ and any other $j$-th pseudo-class $\mathbf{\hat{q}_{j}}$~(including $\mathbf{\hat{q}^{a}_{j}}$ and $\mathbf{\hat{q}^{b}_{j}}$) should be maximized. 
Consequently, different from single-modality clustering in ~\cite{li2021contrastive}, for the $i$-th pseudo-class result $\mathbf{\hat{q}^{a}_{i}}$, we obtain the positive pseudo-class set $\mathbf{\hat{P}^a_i}=\{\mathbf{\hat{q}^{b}_{i}\}}$, and the negative pseudo-class set $\mathbf{\hat{N}^{a}_i}=\{\mathbf{\hat{q}^{a}_{j}}, \mathbf{\hat{q}^{b}_{j}} | \mathbf{\hat{q}^a}_j ,  \mathbf{\hat{q}^b}_j \in \mathcal{B}_{\hat{q}}, j \neq i\}$ containing $2N_{cls}-2$ different classes from both intra-modality and inter-modality.
Thus, the pseudo-classification aligning loss for $\mathbf{\hat{q}^{a}_{i}}$ can be expressed as
\begin{equation}
\begin{split}
\hat{L}^{({a})}_{i} = -\log\frac{\exp(\mathbf{\hat{q}^{a}_{i}} \cdot \mathbf{\hat{q}^{b}_{i}}/\hat{\tau})}{\exp(\mathbf{\hat{q}^{a}_{i}} \cdot \mathbf{\hat{q}^{b}_{i}}/\hat{\tau}) + \sum_{\mathbf{\hat{q}_{j}} \in \mathbf{\hat{N}^{a}_i}}\exp(\mathbf{\hat{q}^{a}_{i}} \cdot \mathbf{\hat{q}_{j}}/\hat{\tau})}.
\label{eq:pcl}
\end{split}
\end{equation}
Classification labels are applied across all modalities in an equal way. Thus, different from the cross-modal feature contrastive loss, $\hat{L}^{({a})}_{i}$ and $\hat{L}^{({b})}_{i}$ are symmetric.
The multimodal pseudo-classification aligning loss $L_{\text{MPC}}$ for the mini-batch $\mathcal{B}$ of $N$ input paired samples with $N_{cls}$-way pseudo-classification can be expressed as
\begin{equation}
L_{\text{MPC}}=\frac{1}{2N_{cls}}\sum^{N_{cls}}_{i=1}(\hat{L}^{({a})}_{i} + \hat{L}^{({b})}_{i} ) + \lambda_{\text{PR}} L_{\text{PR}},
\label{eq:MPC}
\end{equation}
where $L_{\text{PR}}=-\sum_{i=1}^{N_{cls}}P(\mathbf{\hat{q}{i}}) \log P(\mathbf{\hat{q}{i}})$ is a Shannon entropy-based regularization loss~\cite{li2021contrastive} with $\lambda_{\text{PR}} \in \mathbb{R}^-$ acting as a scale weight. It is employed to prevent the situation where most samples are classified into the same pseudo-class.

\input{insert_figures/res_design}
\subsubsection{Multi-task combination.} After getting $L_{\text{CMF}}$ from cross-modal feature contrastive learning and $L_{\text{MPC}}$ from multimodal pseudo-classification aligning, we combine them to obtain the pre-training loss $L_{\text{PT}}$ as
\begin{equation}
\begin{split}
L_{\text{PT}}=L_{\text{CMF}}+ \delta L_{\text{MPC}},
\label{eq:loss_ft}
\end{split}
\end{equation}
where $\delta \in \mathbb{R}^+$ is computed by the values of $L_{\text{CMF}}$ and $L_{\text{MPC}}$ within each mini-batch for loss balancing.

Figure \ref{fig:res_design} demonstrates the contrastive feature space and the pseudo-classification space after MESEN's multimodal-aided pre-training stage, demonstrating its ability to maintain distinct modality spaces and ensure multimodal pseudo-classification alignment.

\subsection{Unimodal Fine-tuning Stage}
During fine-tuning, we obtain the pre-trained unimodal encoder according to the specific modality used in subsequent unimodal HAR, and refine it with a classifier head using labeled data.
However, this process involves two potential issues, i.e., loss of knowledge from pre-training and overfitting due to label scarcity, both of which will degrade recognition performance.

To mitigate these issues, we employ a layer-aware fine-tuning mechanism with the regularization loss $L_{\text{FR}}$.
During fine-tuning, the model consists of two parts: the pre-trained encoder which is to be fine-tuned, and the classifier head which needs to be trained from scratch.
Their parameters are denoted by $\theta_{e}$ and $\theta_{c}$, respectively.
The regularization loss $L_{\text{FR}}$ can be expressed as
\begin{equation}
\begin{split}
L_{\text{FR}}= \sum_{i=1}^{n_e}(\gamma_{i}||{\theta_{e}}_{i}||^2)+||\theta_{c}||^2,
\end{split}
\label{eq:loss_fr}
\end{equation}
where $i$ denotes different layers of the encoder and $n_e$ is the total layer number of the encoder, $\gamma_{i}$ controls the degree to which the learned knowledge from the pre-training stage is retained.
If the equal weight $\gamma_{i}=1$ is assigned to all the layers, the loss becomes $L_{\text{FR}}=||\theta_{e}||^2 + ||\theta_{c}||^2$.
The unimodal fine-tuning loss $L_{\text{FT}}$ can be expressed as 
\begin{equation}
\begin{split}
L_{\text{FT}}=L_{\text{CLS}}+\lambda_{\text{FR}} L_{\text{FR}},
\label{eq:loss_ft}
\end{split}
\end{equation}
where $L_{\text{CLS}}$ represents the classification loss, and $\lambda_{\text{FR}}$ acts as a scale weight for the regularization term.

%% file: insert_figures/framework.tex
\begin{figure}[t]
    \centering
    \captionsetup{skip=8pt}
    \includegraphics[width=0.49\textwidth]{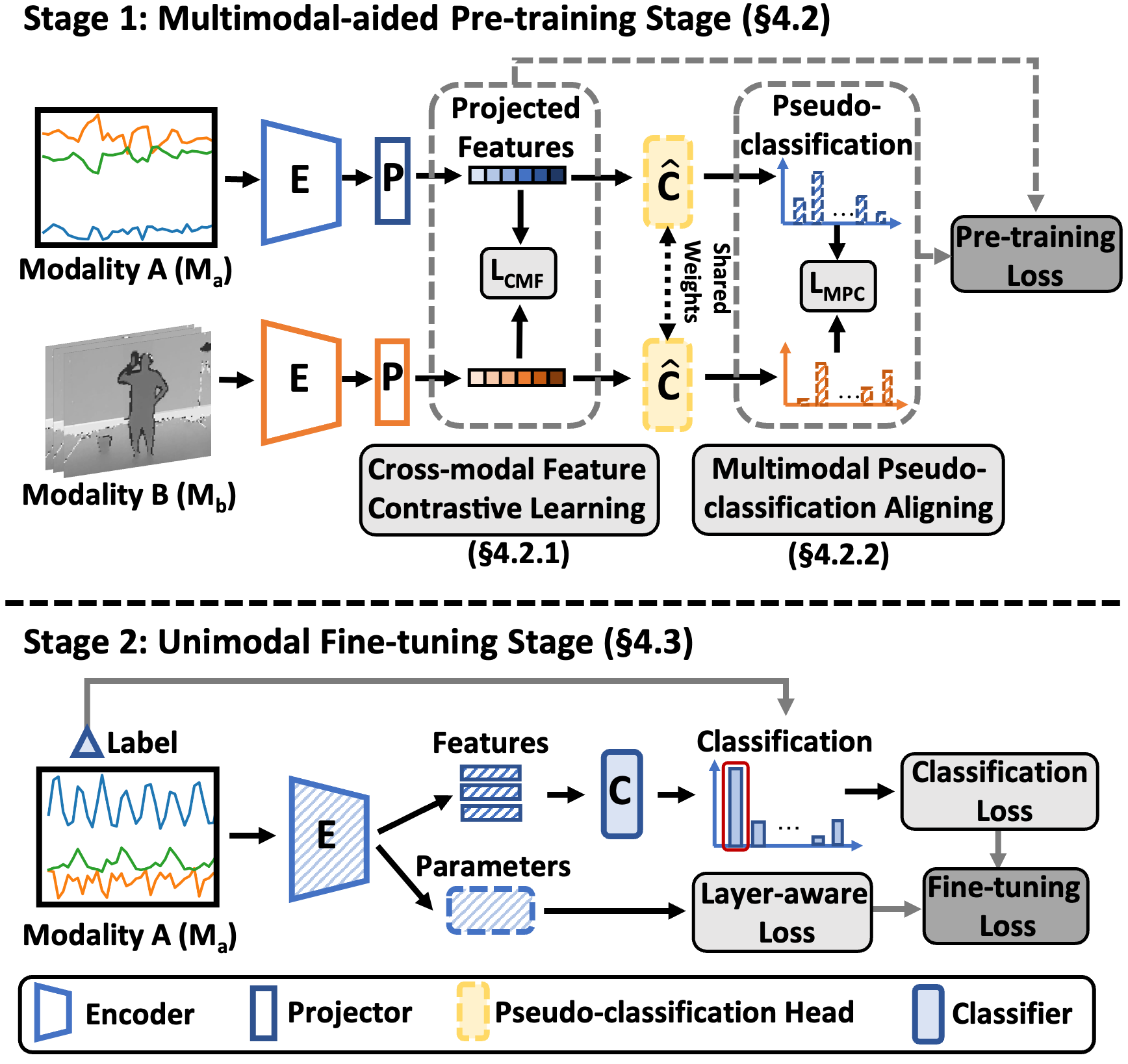}
    \caption{The overview of MESEN.}
    \label{fig:design}
\end{figure}

%% file: insert_figures/modality_consistent.tex
\begin{figure}[t]
    \centering
    \captionsetup{skip=8pt}
    \includegraphics[width=0.48\textwidth]{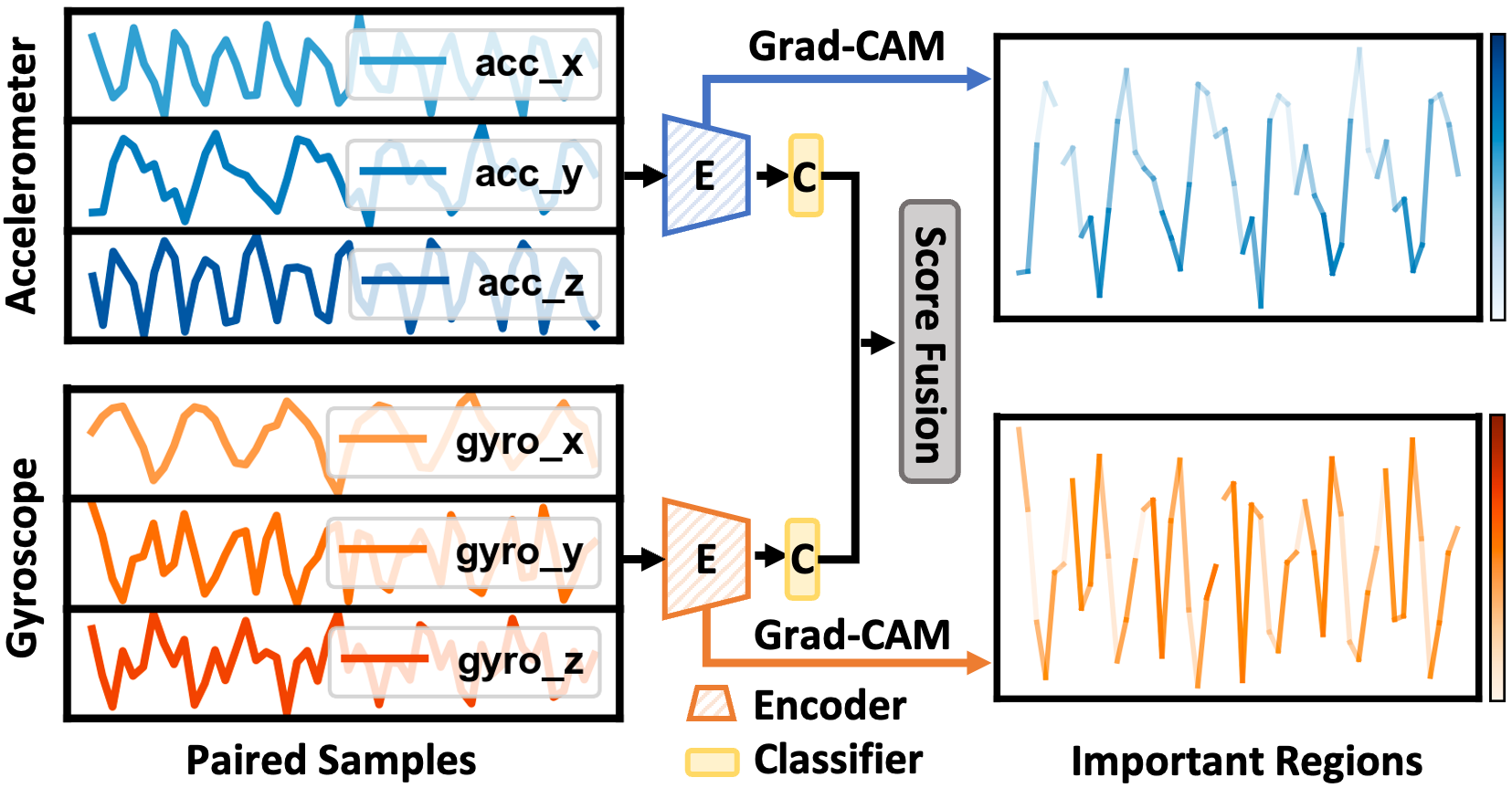}
    \caption{Temporally aligned paired samples of two modalities, recording the same activity~(\textit{Walking}), exhibit similar variation patterns and correlated important regions. The color intensity indicates the importance of the region.}
    \label{fig:modality_consistent}
\end{figure}

%% file: insert_figures/feature_space.tex
\begin{figure}[t]
    \centering
    \captionsetup{skip=8pt}
    \includegraphics[width=0.48\textwidth]{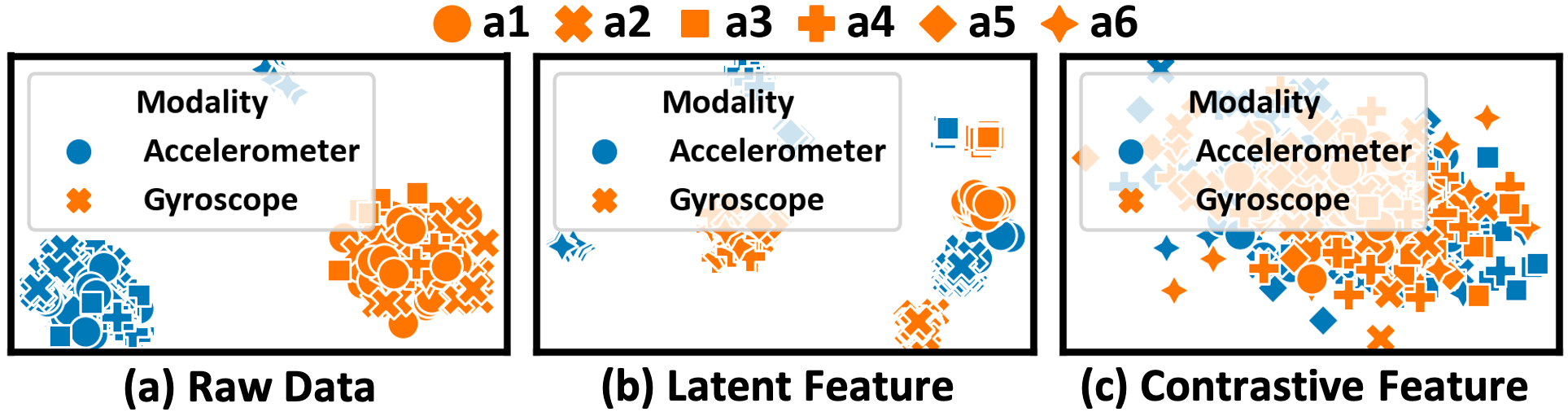}
    \caption{{\textbf{(a) \& (b)}:} Distinct modality spaces of different modality data are clear both at the raw data level and the latent feature representation level.
    {\textbf{(c):}} modality spaces are distorted in contrastive learning. 
    }
    \label{fig:feature_space}
\end{figure}

%% file: insert_figures/modalitycontrastive.tex
\begin{figure}[t]
    \centering
    \captionsetup{skip=8pt}
    \includegraphics[width=0.46\textwidth]{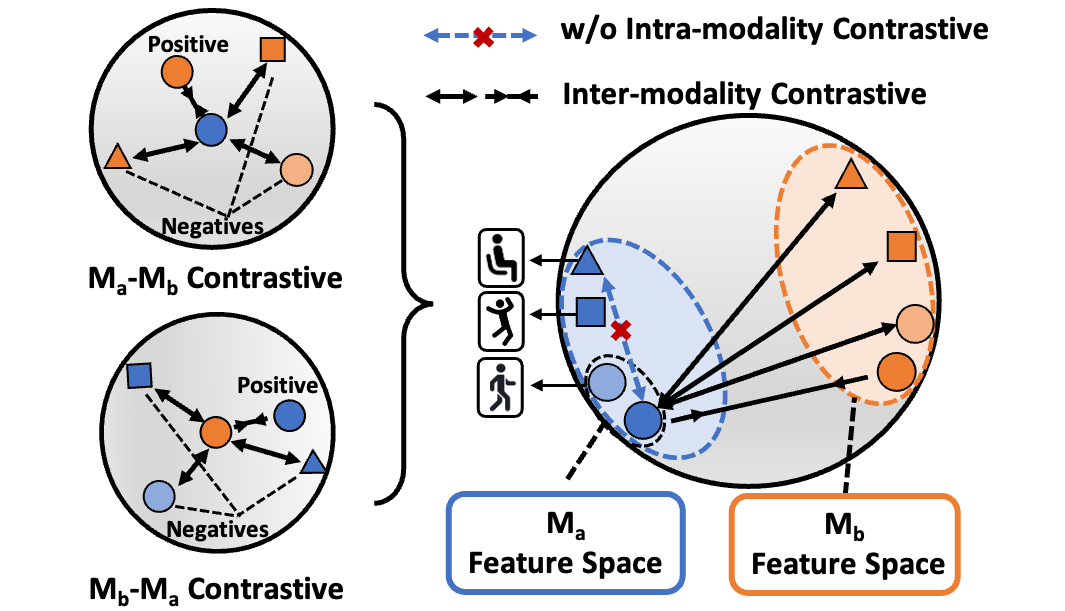}
    \caption{The design of cross-modal feature contrastive learning. }
    \label{fig:cmcl}
\end{figure}

%% file: insert_figures/decision_cls_space.tex
\begin{figure}[t]
    \centering
    \captionsetup{skip=8pt}
    \includegraphics[width=0.47\textwidth]{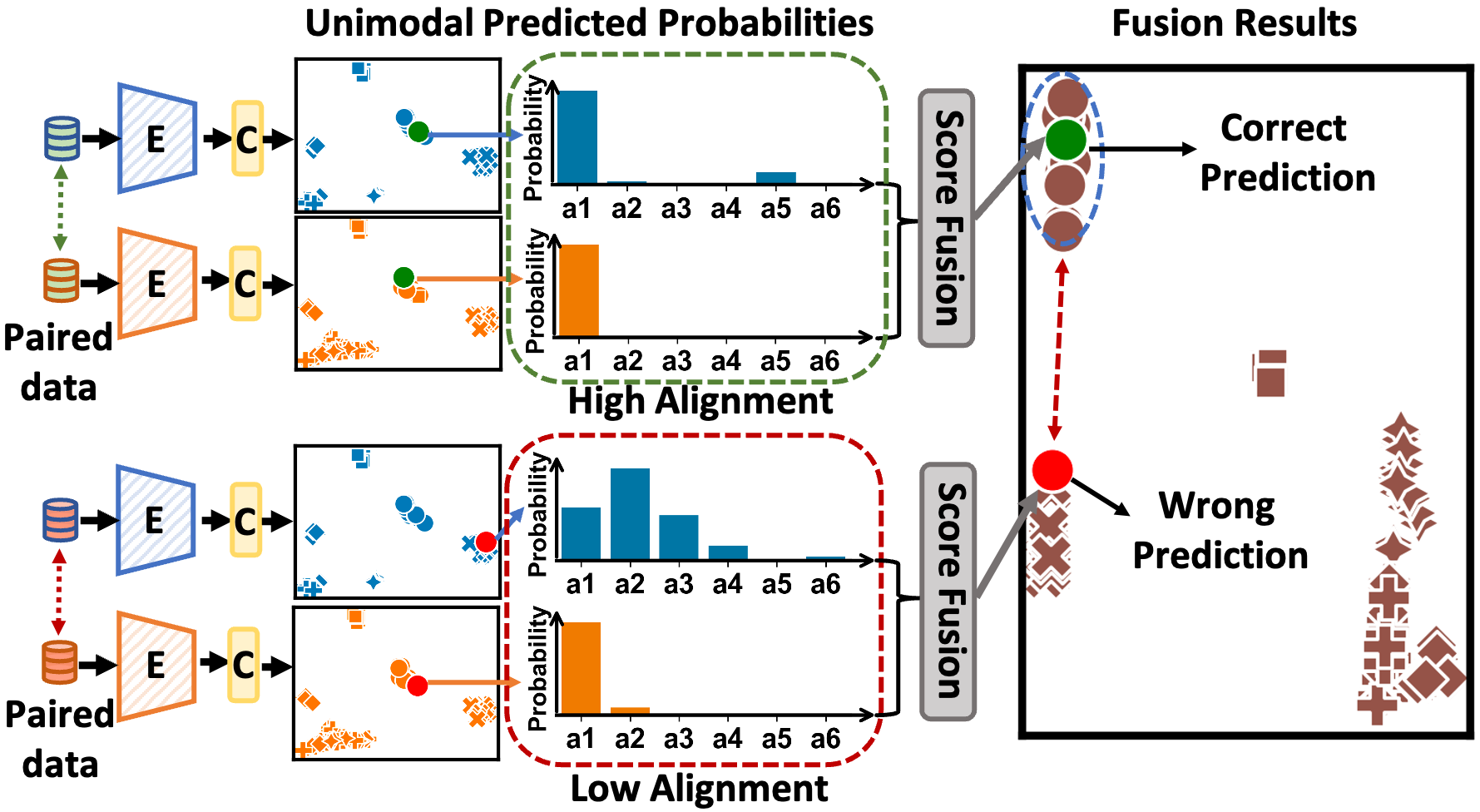}
    \caption{The effects of alignment between paired unimodal predicted probabilities on final fusion results. 
       }
    \label{fig:decision_cls_space}
\end{figure}

%% file: insert_figures/classificationcontrastive.tex
\begin{figure}[t]
    \centering
    \captionsetup{skip=8pt}
    \includegraphics[width=0.46\textwidth]{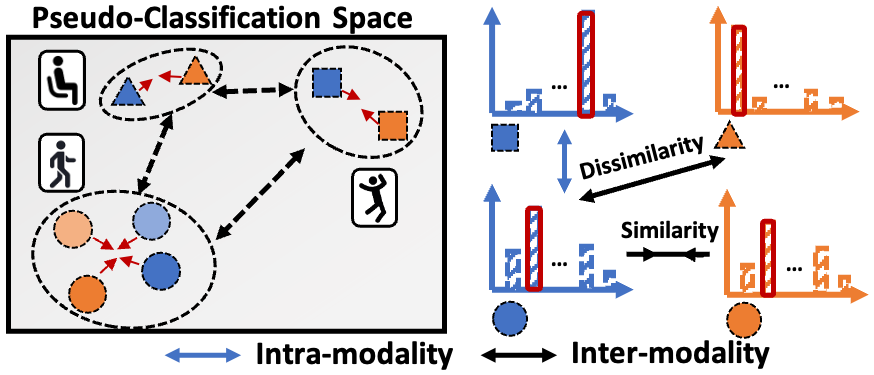}
    \caption{The design of multimodal pseudo-classification aligning.}
    \label{fig:pccl}
\end{figure}

%% file: insert_figures/res_design.tex
\begin{figure}[t]
    \centering
    \captionsetup{skip=8pt}
    \includegraphics[width=0.48\textwidth]{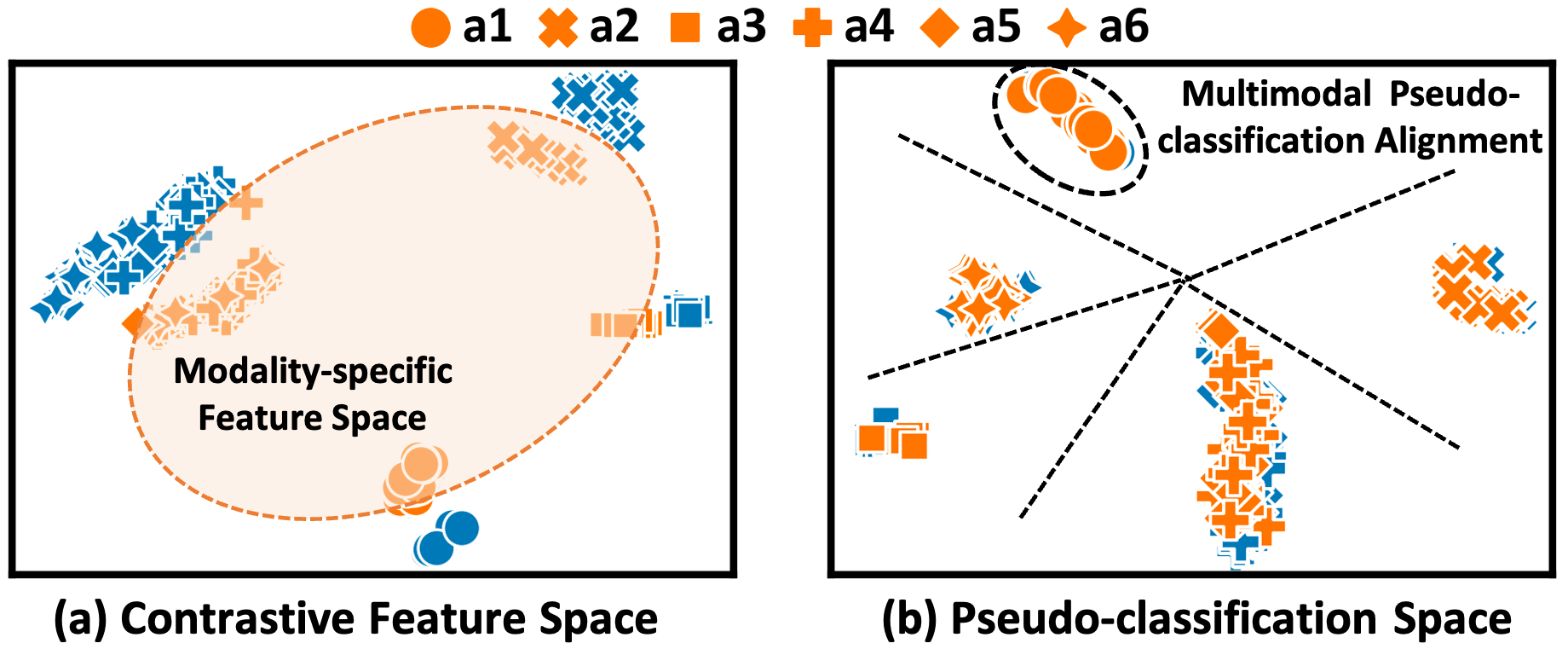}
    \caption{The performance of MESEN design.}
    \label{fig:res_design}
\end{figure}

%% file: 7_Evaluation.tex
\section{Performance Evaluation}
\label{sec:evaluation}

\subsection{Datasets and Methodology}
\label{sec:dataset}
\subsubsection{Multimodal HAR Datasets.} 
To evaluate the effectiveness of MESEN, we use eight multimodal datasets that span diverse modalities~(accelerometer, gyroscope, magnetometer, skeleton points, depth images, and mmWave radar), activities, user scales, and collection environments. 
Table \ref{table:dataset} provides a summary of these datasets.

    \noindent\textbf{UCI dataset~\cite{reyes2016transition}.} It contains accelerometer and gyroscope data from 30 users performing 6 activities with a smartphone (Samsung Galaxy S II) on the user's waist. The sampling rate is $50\,\text{Hz}$.
    
    \noindent\textbf{MotionSense dataset~\cite{malekzadeh2019mobile}.} It comprises data collected by accelerometer and gyroscope sensors with a sampling rate of $50\,\text{Hz}$. During data collection, a total of 24 users performed 6 activities with an iPhone 6s placed in the user's front pocket in the same environment and conditions.
    
    \noindent\textbf{HHAR dataset~\cite{stisen2015smart}.} It contains accelerometer and gyroscope readings collected by a variety of smartphones from 9 users performing 6 daily activities.
    During data collection, the devices were carried by the users around their waists with the sampling rate ranging from $100\,\text{Hz}$ to $200\,\text{Hz}$.
    
    \noindent\textbf{USC dataset~\cite{zhang2012usc}.} It consists of accelerometer and gyroscope data from 14 users performing 12 activities based on their own style. During data collection, the device was placed at the user's front right hip with a reachable sampling rate of $100\,\text{Hz}$.
    
    \noindent\textbf{Shoaib dataset~\cite{shoaib2014fusion}.} It contains magnetometer readings along with accelerometer and gyroscope data collected by Samsung Galaxy SII smartphones from 10 users performing 7 activities.
    The sampling rate is $50\,\text{Hz}$.
    During data collection, devices were placed in five different positions~(\textit{right pocket}, \textit{left pocket}, \textit{belt}, \textit{upper arm}, and \textit{wrist}) on users.

    \noindent\textbf{Cosmo-MHAD dataset~\cite{ouyang2022cosmo}.} It includes multimodal snippets from 30 users freely performing 14 activities. The dataset contains data of three modalities (depth, IMU, and mmWave radar) collected at the sampling rate of $20\,\text{Hz}$, $100\,\text{Hz}$, and $15\,\text{Hz}$, respectively. Due to the poor performance of radar when used alone as reported in ~\cite{ouyang2022cosmo}, we use only IMU and depth data from the dataset. 
    
    \noindent\textbf{mRI dataset~\cite{an2022mri}.} It is a multimodal 3D human pose dataset. The released mRI includes over 5 million frames of mmWave and IMU data from 20 users with a sampling rate of $10\,\text{Hz}$ and $50\,\text{Hz}$, respectively. The IMU data are from 6 sensors placed on different positions~(\textit{left wrist}, \textit{right wrist}, \textit{left knee}, \textit{right knee}, \textit{head}, and \textit{pelvis}) of the user. 
    Applying a 1-second window on the frame sequences, we obtained 4,105 samples of 11 movements from the dataset.

    \noindent\textbf{UTD dataset~\cite{chen2015utd}.} It contains data collected by a Microsoft Kinect sensor and a wearable inertial sensor with a sampling rate of $30\,\text{Hz}$ and $50\,\text{Hz}$, respectively. We use IMU, skeleton, and depth modalities from it. During data collection, 8 subjects performed 27 different actions in an indoor environment. The inertial sensor was placed on the wrist or the thigh depending on the type of action.

\input{tables/1_dataset.tex}
For evaluation, we split users into two different subsets for training and inference~(including validation and testing), respectively.
The specific details of the user subsets are demonstrated in Table \ref{table:dataset}.
Moreover, the data in the inference subset is evenly divided into validation and testing sets, maintaining a 1:1 ratio.

\subsubsection{Baselines.} MESEN is evaluated and compared with baselines covering three aspects: supervised unimodal performance, self-supervised unimodal performance, and multi-to-unimodal~(\textit{m2u}) performance which aligns with our training mode.
The baselines used for comparison are as follows.

\noindent\textbf{\textit{Labeltrain}.} It is the supervised learning method that uses labeled data of every single modality in the datasets to predict activities, serving as the supervised unimodal baseline.

\noindent\textbf{SimCLR~\cite{chen2020simple} and CC~\cite{li2021contrastive}.} These two methods are state-of-the-art contrastive-based approaches in computer vision tasks, creating two distinct views of the same sample through data augmentation for conducting single-modality contrastive learning.
We implement them for each modality in the multimodal datasets, serving as the self-supervised unimodal baselines. 

\noindent\textbf{CPCHAR~\cite{haresamudram2021contrastive}.} It is a state-of-the-art self-supervised learning method designed for IMU data. The method utilizes the temporal structure of IMU data. We implement it for the IMU modality.

\noindent\textbf{CMC~(m2u)~\cite{tian2020contrastive}.} It is a leading multi-view contrastive learning method in computer vision tasks, training the modality encoders by directly contrasting multimodal features.
In our experiments, considering the limitation of only unimodal data available during the deployment phase in our target scenarios, we implement CMC under the \textit{m2u} mode, pre-training the modality encoders with multimodal data and fine-tuning each encoder using unimodal data individually as shown in Figure \ref{fig:framework_compare} (c).

\noindent\textbf{Cosmo~(m2u)~\cite{ouyang2022cosmo}.} It is a state-of-the-art contrastive-based multimodal HAR method.
It features a feature fusion contrastive learning approach for extracting effective information from unlabeled multimodal data.
We implement it under the \textit{m2u} mode, maintaining all other parts as in \cite{ouyang2022cosmo}, except for the multimodal feature fusion during fine-tuning.

These baselines can be categorized as the supervised unimodal learning baseline and the contrastive learning baselines.

\input{tables/2_overall_performance_7}
\subsubsection{Configurations.}
MESEN and other baseline models are implemented by using Python and Pytorch~\cite{paszke2019pytorch}.
They are implemented in a server with 4 NVIDIA GeForce RTX 3090 GPUs, 96 GB memory, and an Intel(R) Xeon(R) Gold 6326~(2.90GHz) CPU.
Besides, we also run the unimodal fine-tuning stage on Jetson Nano~\cite{nano} to show the performance of MESEN on edge nodes.

To perform a fair comparison, we conduct baselines and MESEN with the same modality encoders and classifier heads under the same experiment settings, including the same hyperparameters and the same dataset split.
The modality encoders consisting of convolutional layers and transformer encoder layers~\cite{vaswani2017attention} are utilized for all modalities except for the skeleton modality.
We use co-occurrence~\cite{li2018co} as the modality encoder for skeleton data.
The modality projectors used during pre-training are two-layer convolutional layers.
The classifier heads are single linear layer classifiers with softmax activation. 
The learning rate is set at 0.001 for both pre-training and supervised training.
The batch size of the pre-training stage is set at 128.
For supervised training with labeled data, the batch size is generally set at 64. If the number of labels utilized is smaller than 64, the batch size corresponds to the exact number of labels.
Each experiment is conducted independently five times, with different model initialization for each repetition, to mitigate the effect of model initialization on performance.

\subsubsection{Metrics.} We employ both accuracy and F1-score to measure the performance of baselines and MESEN. 
Accuracy measures the proportion of correctly predicted samples among all samples. F1-score considers both false positives and false negatives for each activity category.

\subsection{Evaluation Results}
\subsubsection{Overall performance.} 
\label{sec:overall_performance}

Labeled data are usually scarce in real-world HAR applications.
To evaluate the effectiveness of MESEN, we utilize only a few labeled samples during training.
Table \ref{table:overall_2} provides the performance comparison of MESEN and other baselines under the situations, where only one labeled sample per activity is utilized during training.
The labeling rate is defined as the ratio of labeled samples within the whole training set. 
For example, we employ six labeled samples in total for six activities on the UCI dataset, amounting to 0.35\% of the training set.

According to the results, \textit{Labeltrain} exhibits poor performance on all datasets due to label scarcity, while other methods achieve comparatively better performance by utilizing both labeled data and available unlabeled unimodal or multimodal data.
To be specific, SimCLR and CC perform better than \textit{Labeltrain} by utilizing unlabeled unimodal data. 
However, their performance still falls short of MESEN as they do not fully exploit the available multimodal data.
CMC~(m2u) and Cosmo~(m2u) achieve performance improvements for datasets with minor modality gaps, such as the USC dataset. 
However, they struggle with datasets containing significant heterogeneous modalities like the UTD dataset, where they fail to enhance the recognition performance for all modalities and may even have a negative impact on the performance.

MESEN achieves state-of-the-art results across all datasets, demonstrating its effectiveness in enhancing unimodal HAR performance by exploiting available unlabeled multimodal during training. 
Relative to the top-performing baselines for each modality, MESEN obtains \textbf{5.5\% - 45.2\%} accuracy improvements, with most gains exceeding 10\%.
On average, the recognition accuracy of MESEN on all datasets is 65.7\%, notably outperforming other methods~(\textbf{30.7\%}, \textbf{25.2\%}, \textbf{27.0\%}, \textbf{25.8\%}, and \textbf{27.8\%} higher than \textit{Labeltrain}, SimCLR, CC, CMC~(m2u), and Cosmo~(m2u), respectively).
Moreover, MESEN notably boosts the average F1-score to 63.0\%~(\textbf{34.5\%}, \textbf{26.4\%}, \textbf{28.7\%}, \textbf{26.5\%}, and \textbf{29.9\%} higher than \textit{Labeltrain}, SimCLR, CC, CMC~(m2u), and Cosmo~(m2u), respectively).

\input{insert_figures/nshot}
\input{insert_figures/pretrain_num}
\subsubsection{Impact of labeled sample size.}
We evaluate MESEN and the baselines with different numbers of labeled samples to understand the impact of the labeled sample scale. 
Specifically, we adopt the settings of utilizing $n$ labeled samples per activity during training.
With different $n$, we have labeling rates that range from 0.06\% to 21.32\% on all the datasets. 
As shown in Figure \ref{fig:nshot}, all methods exhibit performance improvements as the number of labeled samples increases.
On one hand, MESEN consistently outperforms the baselines under all settings of $n$ across all datasets.
On the other hand, MESEN demonstrates a more significant performance improvement when a very small amount of labeled samples is used, indicating its effectiveness in practical scenarios with only a few available labeled samples.
Furthermore, even with limited labeled data (ranging from a labeling rate of 0.12\% to a labeling rate of 21.32\% across different datasets), MESEN achieves performance comparable to or even better than supervised learning with 100\% labeled data. 
This indicates that MESEN effectively extracts effective unimodal features from unlabeled data, reducing the need for extensive labeling.

\input{insert_figures/modality_impact}
\subsubsection{Impact of unlabeled sample size.}
MESEN's key principle is to utilize the increasing availability of unlabeled multimodal data for unimodal HAR enhancement. Therefore, the scale of unlabeled data utilized during multimodal-aided pre-training is crucial for MESEN's performance.
We conduct experiments to show the impact of different amounts of unlabeled multimodal data. 
As demonstrated in Figure \ref{fig:pretrain_num}, we fix the number of labeled samples~(one labeled sample per activity), and gradually increase the volume of unlabeled multimodal data~($\#N$ refers to $N$ times the amount of labeled data).
The performance of MESEN increases with a larger unlabeled data amount, suggesting its ability to extract effective information from available unlabeled multimodal data.

\subsubsection{Impact of modality number and type.}
As shown in Figure \ref{fig:modality_impact}, the performance of MESEN is affected by the number and type of modalities utilized during the multimodal-aided pre-training stage.
On the one hand, MESEN achieves better performance when more modalities are available during pre-training, as multiple modalities provide useful information and guidance for unimodal feature extraction.
On the other hand, the effectiveness of MESEN depends on the individual performance of each modality in the recognition task.
For example, in the UTD dataset, IMU outperforms other modalities when it is used alone. 
Therefore, the performance improvement of the Skeleton modality is more significant when IMU and Skeleton are used for pre-training, compared with the improvement achieved when Depth and Skeleton are used.

\input{insert_figures/universal}
\subsubsection{Impact of different model architectures.} 
MESEN is designed as a universal framework, which is adaptable to various modality encoders and classifiers.
We evaluate the effectiveness of MESEN with varying encoders and classifiers on the UCI dataset. 
We replace the default encoder with 1D-CNN~\cite{tang2020rethinking} and CNN-GRU~\cite{kim2021wearable} individually to conduct experiments with varying encoders.
Similarly, we replace the default classifier with GRU~\cite{cho2014properties} and LSTM~\cite{hochreiter1997long} individually.
As shown in Figure \ref{fig:universal}, the results demonstrate that MESEN achieves performance improvements compared with \textit{Labeltrain}, regardless of the model architecture used during the pre-training and fine-tuning stages.

\input{insert_figures/label_souce}
\subsubsection{Impact of labeled sample source.}
The above experiments are conducted with the user setting as depicted in \S\ref{sec:dataset}. 
In this setting, the data utilized for inference~(validation and testing) comes from users distinct from the user subset involved in the training process.
The user subset used for inference refers to `edge users' and the user subset involved in training is denoted as `other users'.
Furthermore, we evaluate the performance of MESEN when it is fine-tuned with labeled samples from edge users.
Figure \ref{fig:label_source} shows that MESEN can achieve better performance with an average accuracy increase of 4.44\% when the labeled samples used during fine-tuning are from edge users.

\subsection{System Effectiveness}
\input{insert_figures/feature_visualization}
\subsubsection{Feature visualization.}
To demonstrate the benefits of MESEN's multimodal-aided pre-training on unimodal feature extraction, we use t-SNE to visualize the unimodal features extracted by MESEN and \textit{Labeltrain} on the mRI dataset's testing set.
The dataset comprises data from two heterogeneous modalities, IMU and mmWave radar.
As shown in Figure \ref{fig:feature_visualization}, the unimodal features extracted by MESEN after the multimodal-aided pre-training stage demonstrate clear clustering properties, even before the fine-tuning stage with labeled data.
This indicates MESEN's effectiveness in utilizing unlabeled multimodal data during pre-training. 
Consequently, compared with \textit{Labeltrain}, MESEN acquires more effective features for unimodal HAR with few labels, owing to the information learned during pre-training.
\input{insert_figures/val_acc}
\subsubsection{Performance on the edge node.}
We implement the unimodal fine-tuning stage of MESEN on Jetson Nano~\cite{nano} to evaluate its performance on the user edge node.
As shown in Figure \ref{fig:edge_use_all} (a), when fine-tuned with few labels~(one labeled sample per activity) on Jetson Nano, MESEN is more time and energy-efficient than the supervised baseline \textit{Labeltrain}.
Specifically, without incurring any additional memory overhead compared with \textit{Labeltrain}, the fine-tuning stage of MESEN is \textbf{1.67$\times$} and \textbf{1.54$\times$} faster than \textit{Labeltrain} on the UCI dataset for the two modalities~(accelerometer and gyroscope), saving \textbf{27.2\%} and \textbf{32.2\%} energy usage, respectively.

\input{insert_figures/ablation_study}
\subsubsection{Ablation study.}
We evaluate the contributions of three components in MESEN: cross-modal feature contrastive learning~($P_{c1}$), multimodal pseudo-classification aligning ($P_{c2}$), and the layer-aware fine-tuning mechanism ($F_{c3}$).
Figure \ref{fig:ablation_study} shows that combining $P_{c1}$ and $P_{c2}$ significantly improves unimodal HAR performance across various multimodal combinations.
Moreover, Figure \ref{fig:edge_use_all} (b) shows the effectiveness of $F_{c3}$ in mitigating overfitting when MESEN adapts to unimodal HAR with only a few labeled samples.

\subsubsection{Impact of parameters settings.}
We evaluate MESEN's sensitivity to various system settings on the MotionSense dataset.
Figure \ref{fig:micro_benchmark} shows the impact of pre-training batch size, contrastive feature dimension, and the number of pseudo-class on MESEN's performance.
Compared with the other two factors, MESEN is particularly sensitive to the pseudo-class number utilized in multimodal pseudo-classification aligning, which is designed as an effective prompt for recognition as described in \S\ref{sec:multimodal_pca}. 
The difference between the number of pseudo-classes and the actual activity category number can impede feature extraction, thus affecting recognition performance.
However, as the number of activity categories is typically readily accessible and requires no extra effort, the performance decrease can be effectively avoided.

%% file: tables/1_dataset.tex
\begin{table}[t]
\centering
\captionsetup{skip=8pt}
\caption{Dataset summary.}
\resizebox{1\linewidth}{!}{
\begin{tabular}{cccccc}
\toprule[2pt]
\multirow{2}{*}{\textbf{Dataset}} & \multirow{2}{*}{\textbf{Modality}} & \multirow{2}{*}{\textbf{Activity}} & \multicolumn{1}{c}{\textbf{User}} & \multirow{2}{*}{\textbf{Sample}} \\ 
 &  &  & \multicolumn{1}{c}{\textbf{(train/valtest)}} &  \\ 
\midrule[2pt]

UCI~\cite{reyes2016transition} & \multicolumn{1}{c}{Acc, Gyro} & \multicolumn{1}{c}{6} & \multicolumn{1}{c}{30~(24/6)} &  2088 \\ \midrule

MotionSense~\cite{malekzadeh2019mobile} & \multicolumn{1}{c}{Acc, Gyro} & \multicolumn{1}{c}{6} & \multicolumn{1}{c}{24~(19/5)} & 4534  \\ \midrule

HHAR~\cite{stisen2015smart} & \multicolumn{1}{c}{Acc, Gyro} & \multicolumn{1}{c}{6} & \multicolumn{1}{c}{9~(7/2)} & 9166  \\ \midrule
USC~\cite{zhang2012usc} & \multicolumn{1}{c}{Acc, Gyro} & \multicolumn{1}{c}{12} & \multicolumn{1}{c}{14~(10/4)} & 38312\\ \midrule
Shoaib~\cite{shoaib2014fusion} & \multicolumn{1}{c}{Acc, Gyro, Mag} & \multicolumn{1}{c}{7} & \multicolumn{1}{c}{10~(8/2)} & 10500  \\ \midrule

Cosmo-MHAD~\cite{ouyang2022cosmo} & \multicolumn{1}{c}{IMU, Depth} & \multicolumn{1}{c}{14} & \multicolumn{1}{c}{30~(25/5)} & 3434\\ \midrule
mRI\cite{an2022mri} & \multicolumn{1}{c}{IMU, Radar} & \multicolumn{1}{c}{11} & \multicolumn{1}{c}{20~(16/4)} & 4105\\ \midrule
UTD~\cite{chen2015utd} & \multicolumn{1}{c}{IMU, Skeleton, Depth} & \multicolumn{1}{c}{27} & \multicolumn{1}{c}{8~(6/2)} & 861\\ 
 
 \bottomrule[2pt]
\end{tabular}
}
\label{table:dataset}
\end{table}

%% file: tables/2_overall_performance_7.tex
\begin{table*}[t]
    \centering
    \captionsetup{skip=8pt}
    \setlength{\tabcolsep}{3.4pt}
    \caption{Performance comparison. 
    MESEN outperforms baselines on all datasets, with the relative improvements over the best-performing baselines~(marked in \underline{underline}) highlighted in \bluemark{{blue}}.
    }
    \resizebox{1.\linewidth}{!}{
    \begin{tabular}{@{}c cccc cccc cccc cccccc@{}} 
    \toprule[2pt]
    \multirow{1}{*}{\textbf{Dataset}} 
            & \multicolumn{4}{c}{{UCI}} 
            & \multicolumn{4}{c}{{MotionSense}} 
            & \multicolumn{4}{c}{{HHAR}} 
            & \multicolumn{6}{c}{{Shoaib}}\\ 

    \cmidrule(lr){2-5}
    \cmidrule(l){6-9} 
    \cmidrule(l){10-13} 
    \cmidrule(l){14-19}
    
    \multirow{1}{*}{\textbf{Labeling rate}} 
            & \multicolumn{4}{c}{ 0.35\%}   
            & \multicolumn{4}{c}{ 0.17\%}   
            & \multicolumn{4}{c}{ 0.08\%}   
            & \multicolumn{6}{c}{ 0.08\%}   
            \\
    \midrule
    {\textbf{Modality}} 
            & \multicolumn{2}{c}{{Acc}} &  \multicolumn{2}{c}{{Gyro}}   
            & \multicolumn{2}{c}{{Acc}} & \multicolumn{2}{c}{{Gyro}}    
            & \multicolumn{2}{c}{{Acc}} & \multicolumn{2}{c}{{Gyro}}    
            & \multicolumn{2}{c}{{Acc}} &  \multicolumn{2}{c}{{Gyro}}  & \multicolumn{2}{c}{{Mag}} 
            \\
    \midrule

    \multirow{1}{*}{\textbf{Metrics}} 
            & Acc & F1 &  Acc & F1 
            &  Acc & F1 &  Acc & F1 
            &  Acc & F1 &  Acc & F1
            &  Acc & F1 &  Acc & F1 &  Acc & F1\\
    \midrule
    \textit{Labeltrain}  & \num{0.4415} & \num{0.3649 } & \num{0.3532 } & \num{0.2514 }    
                & \num{0.4904 } & \num{0.3913 } & \num{0.4393 } & \num{0.3689 }             
                & \num{0.4457  } & \num{0.3400  } & \num{0.3699 } & \num{0.2784  }        
                & \num{0.3164 } & \num{0.2968 } & \num{0.2867  } & \num{0.1896 }  & \num{0.2269 } & \num{0.1859 }            
                 \\

    SimCLR      & \underline{\num{0.5160  }} & \num{0.4913  } & \underline{\num{0.4596}} & \underline{\num{0.4351  }}        
                & \underline{\num{0.5197  }} & \num{0.4582  } & \num{0.4961  } & \underline{\num{0.4473 } }         
                & \num{0.4875  } & \underline{\num{0.4412 }} & \num{0.4875  } & \num{0.4107  }          
                & \underline{\num{0.3390 } } & \underline{\num{0.3152 }} & \num{0.3333 }  & \num{0.3174  } & \num{0.2537 }   & \num{0.2530 }          
                 \\
                 
    CC          & \num{0.4149  } & \num{0.4249  } & \num{0.4436  } & \num{0.4271  }         
                & \num{0.3013  } & \num{0.2826   } & \num{0.4629 } & \num{0.3695  }         
                & \num{0.4485  } & \num{0.3676 } & \num{0.4345 } & \num{0.3513  }          
                & \num{0.2625 } & \num{0.2497 } & \underline{\num{0.3600 } } & \underline{\num{0.3400 }}  & \num{0.2137 } & \num{0.2124  }             
                 \\
                 
    CMC (m2u)   & \num{0.5053  } & \underline{\num{0.4953 }} & \num{0.3457  } & \num{0.2860 }           
                & \num{0.5070 } & \num{0.4548 } & \underline{\num{0.5061 } } & \num{0.4182  }           
                & \num{0.4507   } & \num{0.4355  } & \num{0.3939  } & \num{0.3533  }       
                & \num{0.3219 } & \num{0.3104 } & \num{0.2943 } & \num{0.2905 }  & \num{0.2962 } & \num{0.2909 }                
                \\

    Cosmo (m2u) & \num{0.5128 } & \num{0.4938 } & \num{0.4404 } & \num{0.3786 }           
                & \num{0.5060  } & \underline{\num{0.4672 } } & \num{0.4932  } & \num{0.4114  }         
                & \underline{\num{0.4886}} & \num{0.4361  } & \underline{\num{0.4932} } & \underline{\num{0.4114 }}         
                & \num{0.3152 } & \num{0.3134 } & \num{0.2981  } & \num{0.2971  }  & \underline{\num{0.3181 }} & \underline{\num{0.3085 }}              
                \\
    \midrule
    \multirow{2}{*}{MESEN~(Ours) }       
    & \textbf{{0.888}} & \textbf{{0.890}} & \textbf{{0.695}} & \textbf{{0.636}}     
                & \textbf{{0.790 }} & \textbf{{0.807}} & \textbf{{0.682 }} & \textbf{{0.694 }}     
                & \textbf{{0.659 }} & \textbf{{0.676 }} & \textbf{{0.660 }} & \textbf{{0.651  }}  
                & \textbf{{0.487 }} & \textbf{{0.445 }} & \textbf{{0.462 }} & \textbf{{0.434 }} & \textbf{{0.458 }} & \textbf{{0.429 }}  \\                   
    
                & \textbf{\blue+{0.372}} & \textbf{\blue+{0.395}} & \textbf{\blue+{0.235}} & \textbf{\blue+{0.201}}     
                & \textbf{\blue+{0.270}} & \textbf{\blue+{0.340}} & \textbf{ \blue+{0.176}} & \textbf{\blue+{0.247}}     
                & \textbf{\blue+{0.170}} & \textbf{\blue+{0.235}} & \textbf{ \blue+{0.167} } & \textbf{ \blue+{0.240} }  
                & \textbf{\blue+{0.148} } & \textbf{\blue+{0.130}} & \textbf{\blue+{0.102} } & \textbf{\blue+{0.094}} & \textbf{\blue+{0.140}} & \textbf{\blue+{0.120} }    
    \\
    \bottomrule[2pt]

    \multirow{1}{*}{\textbf{Dataset}} 
            & \multicolumn{4}{c}{{USC}} 
            & \multicolumn{4}{c}{{mRI}} 
            & \multicolumn{4}{c}{{Cosmo-MHAD}}
            & \multicolumn{6}{c}{{UTD}}\\ 

    \cmidrule(lr){2-5}\cmidrule(l){6-9} \cmidrule(l){10-13} \cmidrule(l){14-19}

    \multirow{1}{*}{\textbf{Labeling rate}} 
            & \multicolumn{4}{c}{ 0.06\%}  
            & \multicolumn{4}{c}{ 0.36\%} 
            & \multicolumn{4}{c}{ 0.51\%} 
            & \multicolumn{6}{c}{ 4.18\%}\\
    \midrule
    \multirow{1}{*}{\textbf{Modality}} 
            & \multicolumn{2}{c}{{Acc}} &  \multicolumn{2}{c}{{Gyro}}  
            & \multicolumn{2}{c}{{IMU}} & \multicolumn{2}{c}{{Radar}} 
            & \multicolumn{2}{c}{{IMU}} & \multicolumn{2}{c}{{Depth}}
            & \multicolumn{2}{c}{{IMU}} & \multicolumn{2}{c}{{Skeleton}} & \multicolumn{2}{c}{{Depth}}\\
    \midrule

    \multirow{1}{*}{\textbf{Metrics}} 
            & Acc & F1 &  Acc & F1 
            &  Acc & F1 &  Acc & F1 
            &  Acc & F1 &  Acc & F1 
            &  Acc & F1 &  Acc & F1 &  Acc & F1\\
    \midrule
    \textit{Labeltrain}  
                & \num{0.2290 } & \num{0.2053  } & \num{0.2390  } & \num{0.1754 }    
                & \num{0.4428 } & \num{0.4028 } & \num{0.2328 } & \num{0.2061 }             
                & \num{0.2883  } & \num{0.2132  } & \num{0.2724 } & \num{0.1869  }        
                & \num{0.4111  } & \num{0.3578 }   & \num{0.6426  } & \num{0.5872 } & \num{0.1759 } & \num{0.1304 }             
                 \\

    SimCLR      & \num{0.3630 } & \num{0.3313  } & \num{0.3170  } & \num{0.2941  }         
                & \num{0.4915  } & \underline{\num{0.4687 }} & \num{0.2716   } & \num{0.2575  }         
                & \num{0.3675 } & \num{0.2883 } & \num{0.2423 } & \num{0.1837  }          
                & \num{0.4519 } & \num{0.4000  }   & \num{0.6704 } & \num{0.6241  } & \num{0.2222 } & \num{0.1706  }             
                 \\
                 
    CC          & \num{0.3260  } & \num{0.2523  } & \num{0.3144 } & \num{0.3020  }         
                & \underline{\num{0.5075  }} & \num{0.4596   } & \underline{\num{0.3731} } & \underline{\num{0.3611 } }         
                & \underline{\num{0.4049 }} & \underline{\num{0.3055} } & \underline{\num{0.3890} } & \num{0.2763 }          
                & \underline{\num{0.4556 } } & \underline{\num{0.4103 } }   & \num{0.6574   } & \num{0.6157   } & \num{0.2037  } & \num{0.1628  }             
                 \\
                 
    CMC (m2u)   & \underline{\num{0.5170}} & \underline{\num{0.4846 }} & \underline{\num{0.4364 }} & \underline{\num{0.4276}}           
                & \num{0.3881 } & \num{0.3660 } & \num{0.2801 } & \num{0.2589  }           
                & \num{0.3282  } & \num{0.2427   } & \num{0.3006  } & \num{0.2605  }       
                & \num{0.3704 } & \num{0.3347 }   & \num{0.6389 } & \num{0.5999 } & \underline{\num{0.3000 }} & \underline{\num{0.2683 } }             
                \\

    Cosmo (m2u) & \num{0.3328 } & \num{0.2712  } & \num{0.2740  } & \num{0.2243 }           
                & \num{0.2015  } & \num{0.1114  } & \num{0.2139 } & \num{0.1969 }         
                & \num{0.2736   } & \num{0.1812   } & \num{0.3552 } & \underline{\num{0.2920} }         
                & \num{0.3407 } & \num{0.2870 }   & \underline{\num{0.6796}} & \underline{\num{0.6353}} & \num{0.2852 } & \num{0.2327  }             
                \\
    \midrule
    \multirow{2}{*}{MESEN~(Ours) }   
                & \textbf{0.723 } & \textbf{0.700 } & \textbf{0.695 } & \textbf{0.684 }     
                & \textbf{0.869 } & \textbf{0.866 } & \textbf{0.825 } & \textbf{0.810 }     
                & \textbf{0.506 } & \textbf{0.392 } & \textbf{0.511 } & \textbf{0.429 }  
                & \textbf{0.628  } & \textbf{0.583 } & \textbf{0.735 } & \textbf{0.702 } & \textbf{0.550 } & \textbf{0.517 }    
                \\
                & \textbf{\blue+{0.206}} & \textbf{\blue+{0.215}} & \textbf{\blue+{0.259}} & \textbf{\blue+{0.256}}     
                & \textbf{\blue+{0.361}} & \textbf{\blue+{0.397}} & \textbf{\blue+{0.452}} & \textbf{\blue+{0.449}}     
                & \textbf{\blue+{0.101}} & \textbf{\blue+{0.086}} & \textbf{\blue+{0.122}} & \textbf{\blue+{0.137} }  
                & \textbf{\blue+{0.172} } & \textbf{\blue+{0.173}} & \textbf{\blue+{0.055}} & \textbf{\blue+{0.067}} & \textbf{\blue+{0.250}} & \textbf{\blue+{0.249}}    
                
    \\
    \bottomrule[2pt]

    
    
    \end{tabular}
    }
    \label{table:overall_2}
    \end{table*}

%% file: insert_figures/nshot.tex
\begin{figure*}[t]
    \captionsetup{skip=5pt}
    \centering
    \includegraphics[width=1.\textwidth]{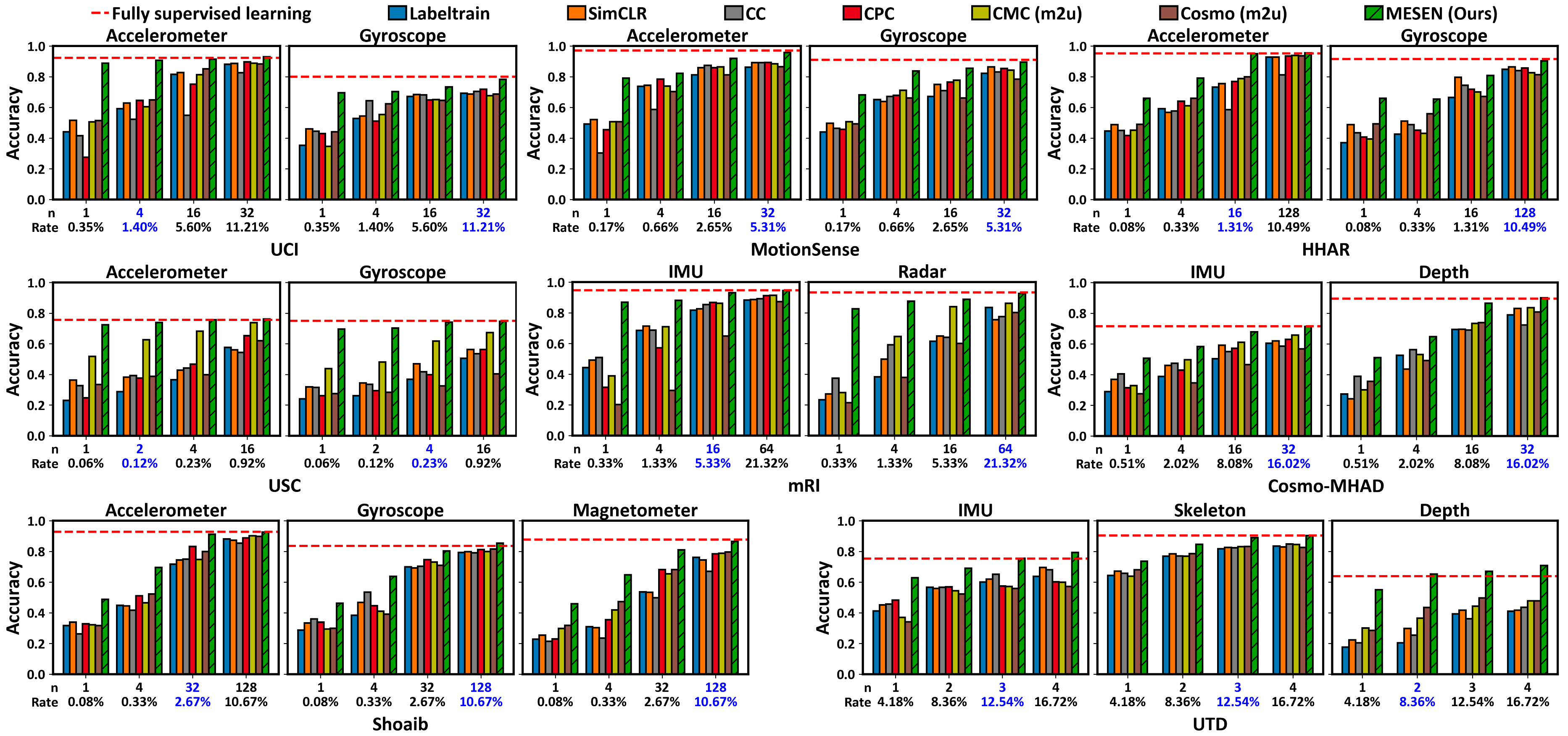}
    \caption{
    The impact of $n$ labeled samples per activity with different $n$ on HAR performance. `Rate' represents the labeling rate. The results of MESEN that are comparable to fully supervised learning with 100\% labeled data are highlighted in \bluemark{{blue}}.}

    \label{fig:nshot}
\end{figure*}

%% file: insert_figures/pretrain_num.tex
\begin{figure}[t]
    \centering
    \captionsetup{skip=8pt}
    \includegraphics[width=0.48\textwidth]{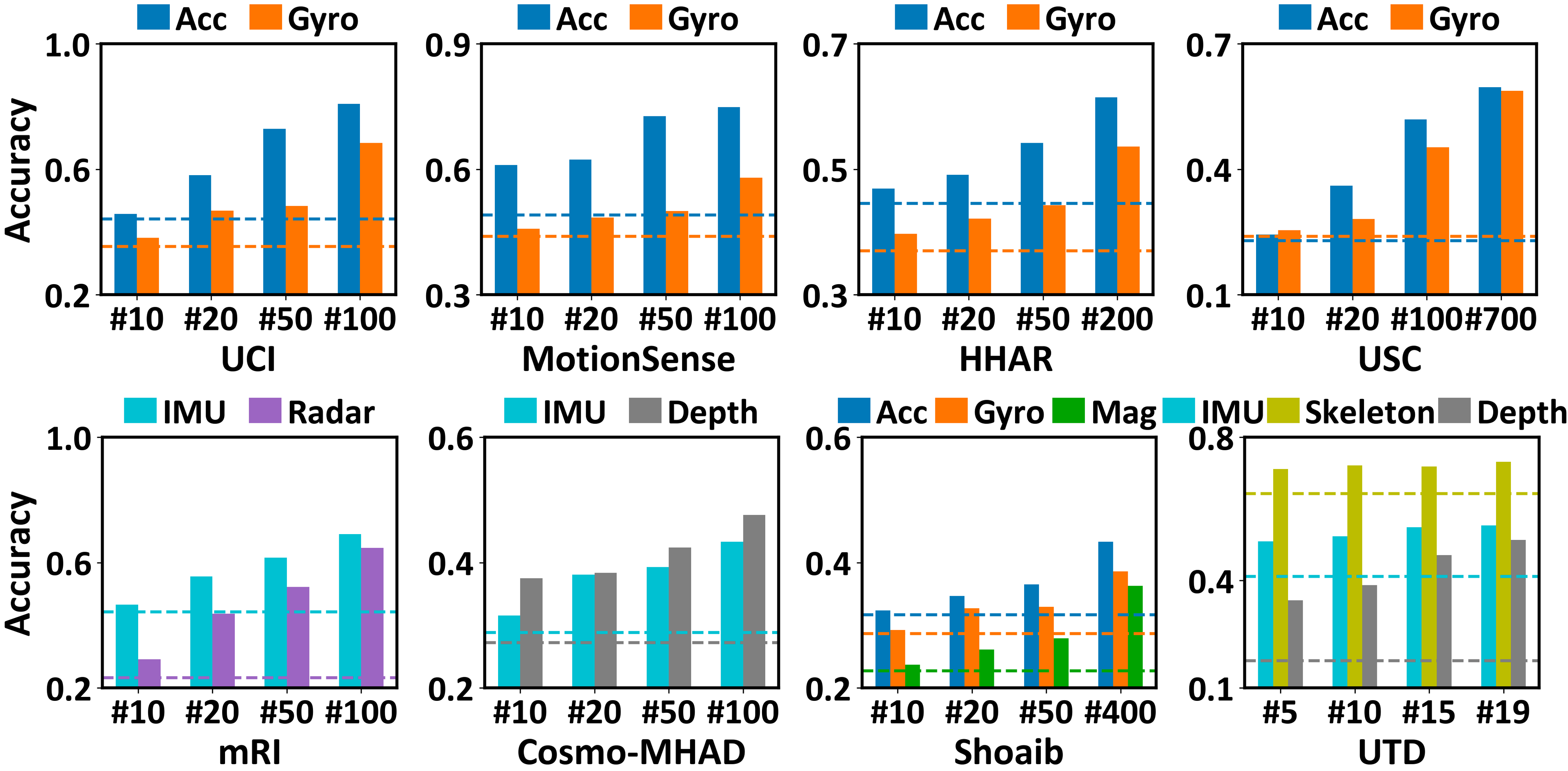}
    \caption{The impact of the available unlabeled multimodal data scale on the performance of MESEN.
The x-axis represents the scale of unlabeled data~($\#N = \frac{\text{Unlabeled data}}{\text{Labeled data}}$), and the dashed lines represent \textit{Labeltrain}'s results.}
    \label{fig:pretrain_num}
\end{figure}

%% file: insert_figures/modality_impact.tex
\begin{figure}[t]
    \centering
    \captionsetup{skip=8pt}
    \includegraphics[width=0.48\textwidth]{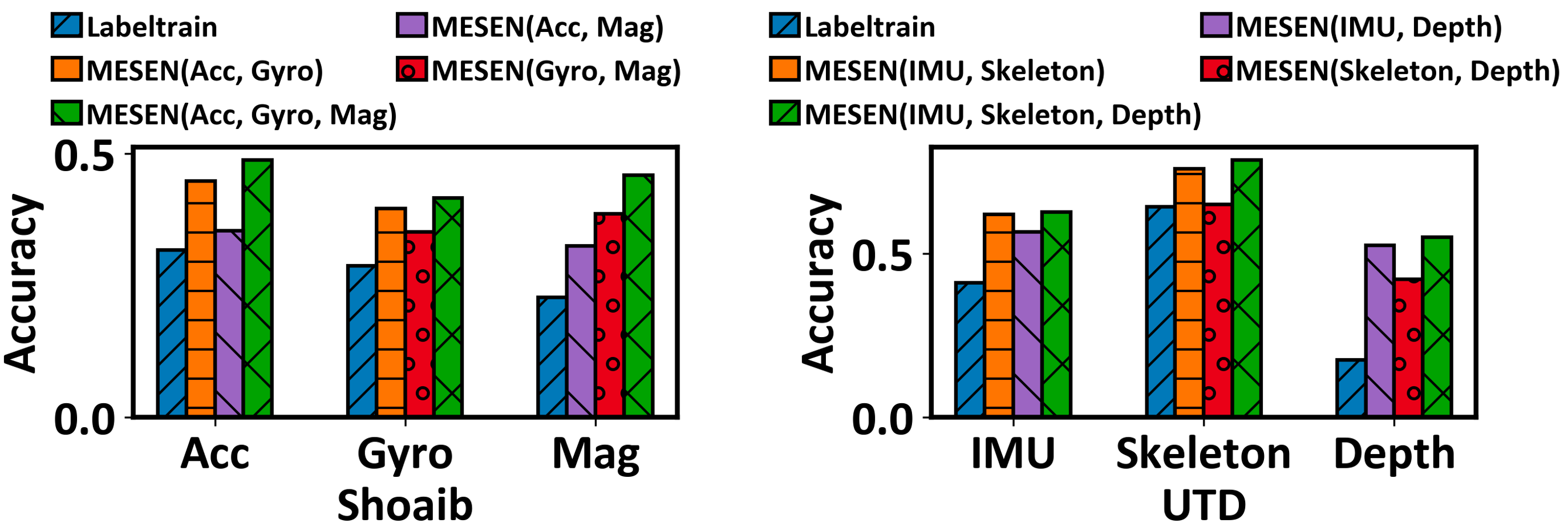}
    \caption{The impact of modality number and type on MESEN's performance. $(\cdot)$ denotes the modalities used during the multimodal-aided pre-training stage.}
    \label{fig:modality_impact}
\end{figure}

%% file: insert_figures/universal.tex
\begin{figure}[t]
    \centering
    \captionsetup{skip=8pt}
    \includegraphics[width=0.47\textwidth]{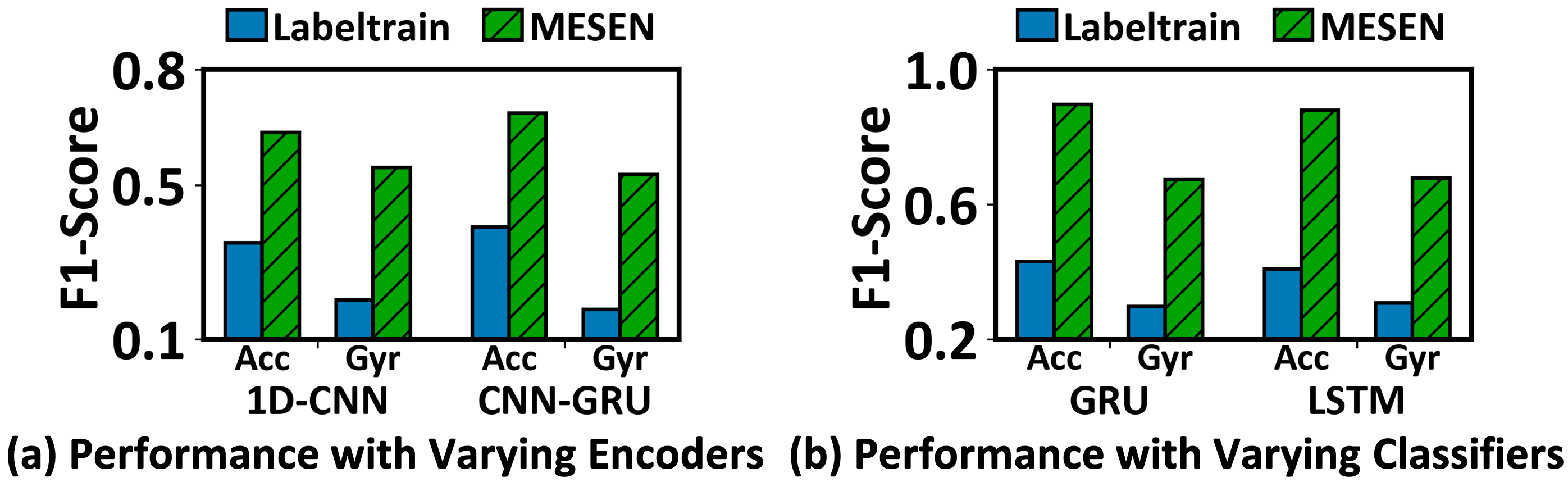}
    \caption{The effectiveness of MESEN with varying encoders and classifiers.}
    \vspace{-0.2em}
    \label{fig:universal}
\end{figure}

%% file: insert_figures/label_souce.tex
\begin{figure}[t]
    \captionsetup{skip=7pt}
    \centering
    \includegraphics[width=0.46\textwidth]{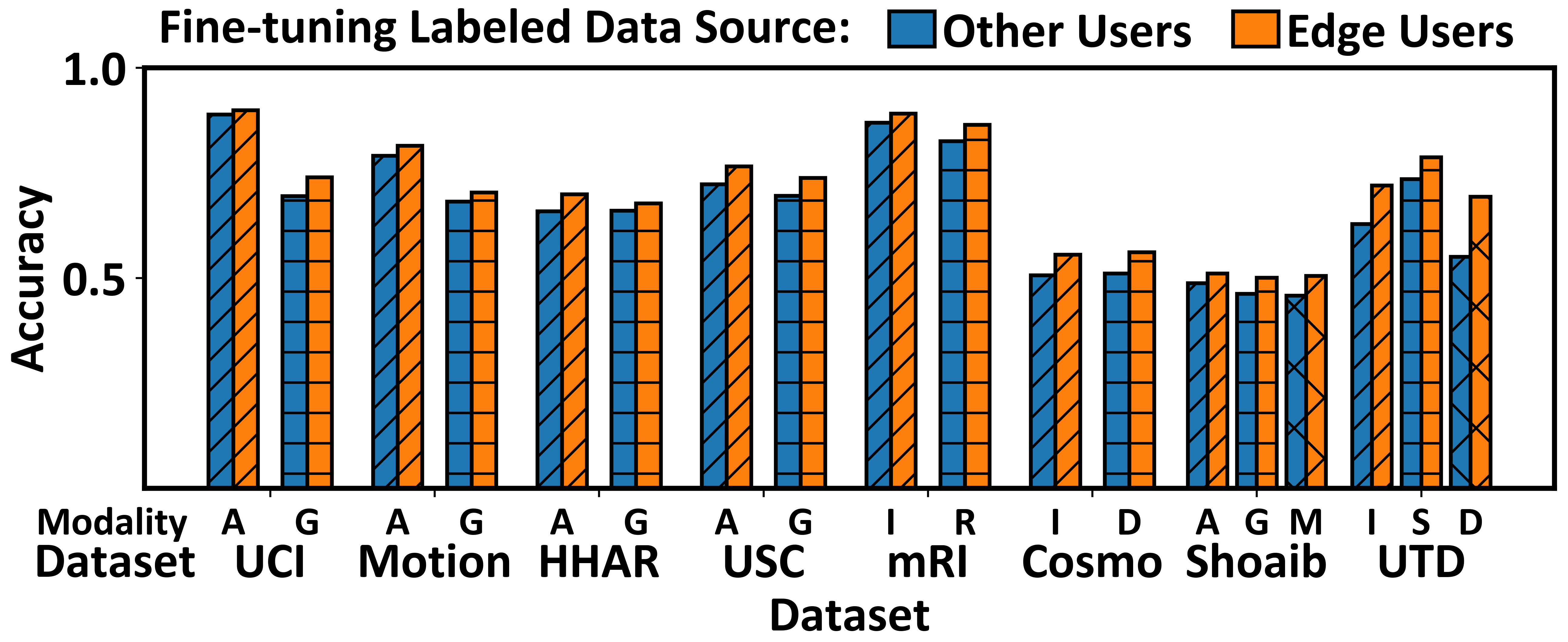}
    \caption{The impact of different labeled data sources on the performance of MESEN.}
    \label{fig:label_source}
\end{figure}

%% file: insert_figures/feature_visualization.tex
\begin{figure}[t]
    \centering
    \captionsetup{skip=8pt}
    \includegraphics[width=0.48\textwidth]{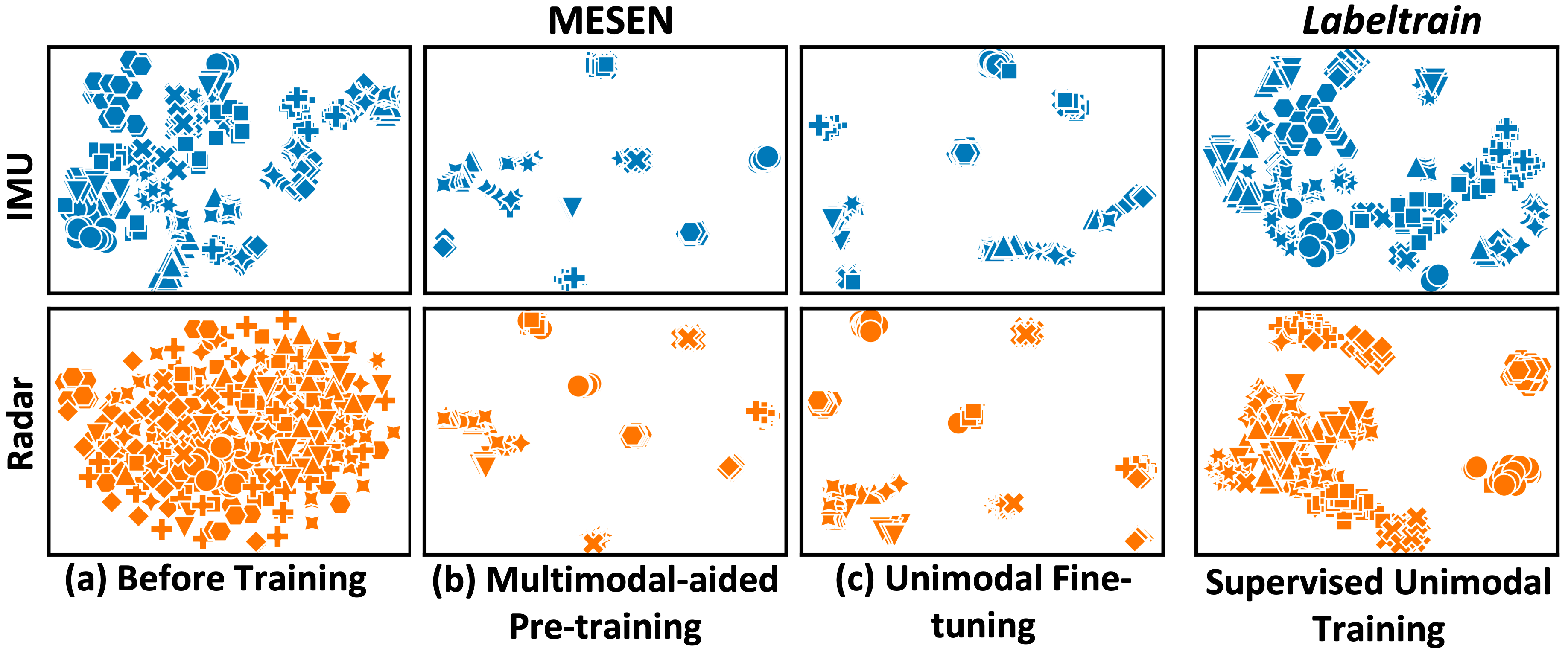}
    \caption{
Unimodal features extracted by MESEN and \textit{Labeltrain}. 
Different shapes denote different activities.}
    \label{fig:feature_visualization}
\end{figure}

%% file: insert_figures/val_acc.tex
\begin{figure}[t]
    \centering
    \captionsetup{skip=8pt}
    \includegraphics[width=0.48\textwidth]{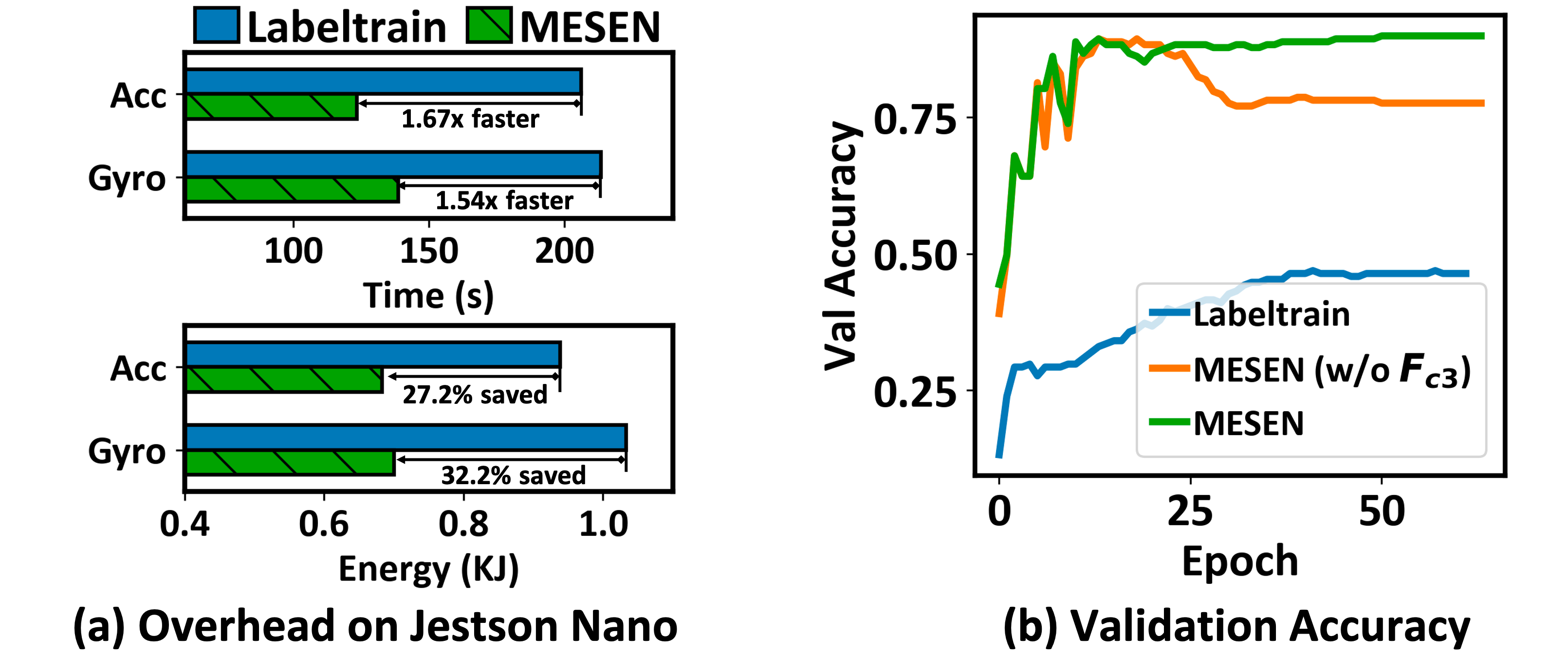}
    \caption{The fine-tuning performance of MESEN.}
    \label{fig:edge_use_all}
\end{figure}

%% file: insert_figures/ablation_study.tex
\begin{figure}[t]
    \centering
    \captionsetup{skip=8pt}
    \includegraphics[width=0.48\textwidth]{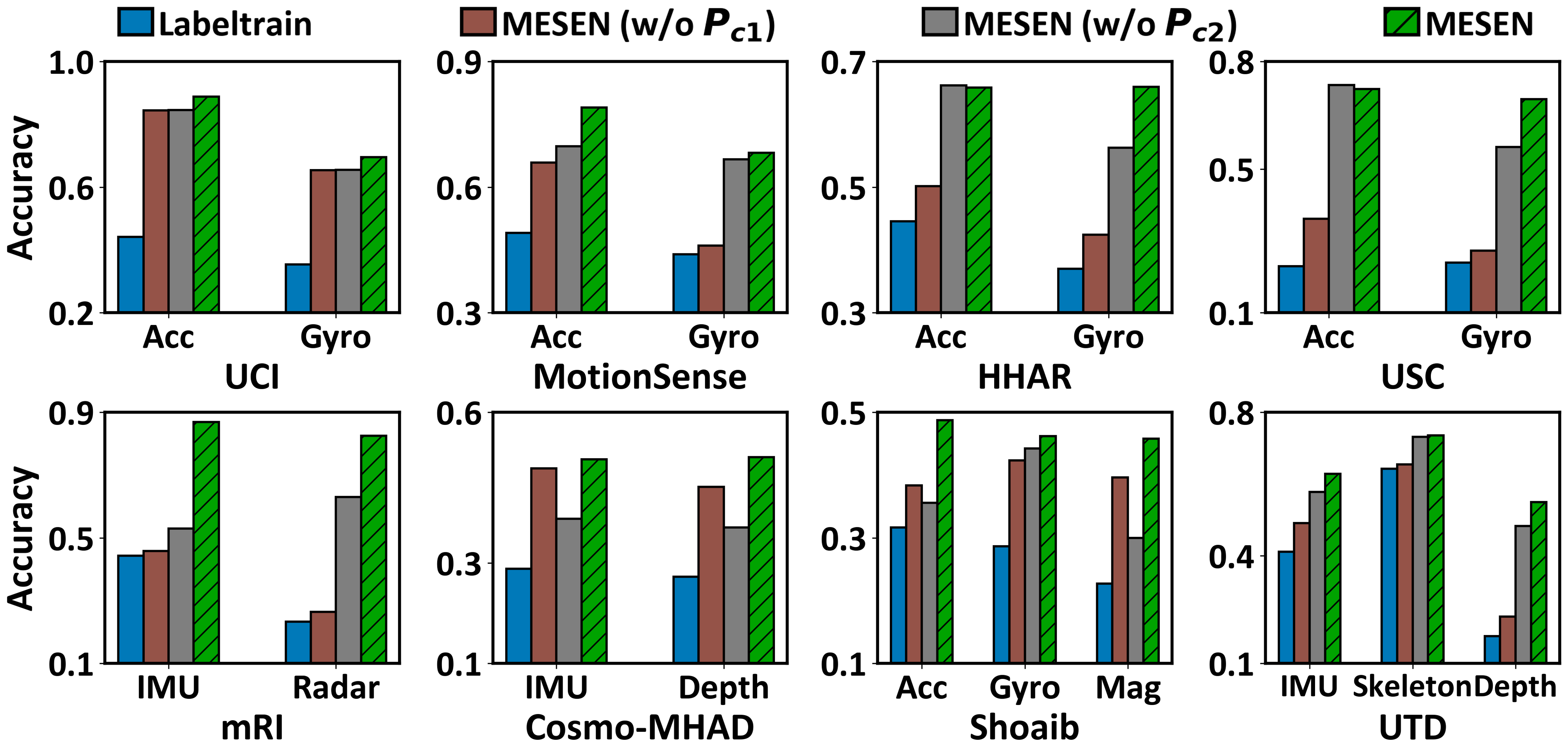}
    \caption{Ablation study for $P_{c1}$ and $P_{c2}$ of MESEN.}
    \label{fig:ablation_study}
\end{figure}

%% file: 8_Discussion.tex
\section{Discussion}
\label{sec:discussion}
\textbf{Scalability.} 
The increasing number of modality pairs will introduce extra training costs during the multimodal-aided pre-training stage.
An additional modality encoder and a projector are needed for computing features of each new modality during pre-training.
Adopting the assumed atomic operation of the contrastive objective function in COCOA~\cite{deldari2022cocoa}, i.e., the dot-product of sample pairs, the time complexity of MESEN is $\mathcal{O}(M^2N^2+M^2{N_{cls}}^2)$, where $M$ is the number of modalities, $N$ is the number of input pairs in the mini-batch $\mathcal{B}$, and $N_{cls}$ is the number of activity categories.
When the activity categories and the number of input pairs are fixed during implementation, the complexity of MESEN will be $\mathcal{O}(M^2)$.
In practical implementation, the extra training costs vary based on the types of additional modalities. For example, compared with the IMU modality, depth images introduce more requirements for computational resources. 

\textbf{Cross-dataset performance.}
To evaluate the performance of MESEN in cross-dataset scenarios, we conduct experiments on the UCI and MotionSense datasets. Following UniHAR~\cite{xu2023practically}, we select four activities~(\textit{still}, \textit{walk}, \textit{walk upstairs}, and \textit{walk downstairs}) contained in both datasets.
For the cross-dataset setting, we train models on the UCI dataset while validating and testing models on the MotionSense dataset. Other experimental settings align with those in \S\ref{sec:overall_performance}.
Under the cross-dataset setting, MESEN achieves accuracy of 0.536 and 0.787 with accelerometer and gyroscope, respectively, while \textit{Labeltrain} exhibits 0.409 and 0.690.
The accuracy gains of 0.127 and 0.097 by MESEN over \textit{Labeltrain} demonstrate its adaptability in cross-dataset scenarios.
However, MESEN is affected by the domain discrepancies between these two datasets, evidenced by its accuracy of 0.970 and 0.994 on the UCI dataset for the four activities.
The \textit{Physics-Informed Data Augmentation} approach proposed by UniHAR~\cite{xu2023practically} to address data heterogeneity provides a solution to improve MESEN in further study.

\input{insert_figures/micro_benchmark}
\textbf{Multimodal inference.} 
MESEN is designed to operate in a multi-to-unimodal mode, which may present limitations in some HAR application scenarios.
When multiple modalities are readily available during the deployment phase, MESEN can utilize these multimodal data streams through slight adaptation. This is because each modality encoder has been effectively trained during MESEN's multimodal-aided pre-training.
For example, with the experimental settings in \S\ref{sec:overall_performance}, MESEN achieves 0.899 accuracy on the UCI dataset by applying multimodal fusion through concatenating multimodal features during fine-tuning and inference.
However, since multimodal fusion is not the primary design focus of MESEN, it might not capture the full potential of such multimodal streams as effectively as Cosmo~\cite{ouyang2022cosmo}.
Further study is needed to extend MESEN to address broader application scenarios.

%% file: insert_figures/micro_benchmark.tex
\begin{figure}[t]
    \centering
    \captionsetup{skip=8pt}
    \includegraphics[width=0.48\textwidth]{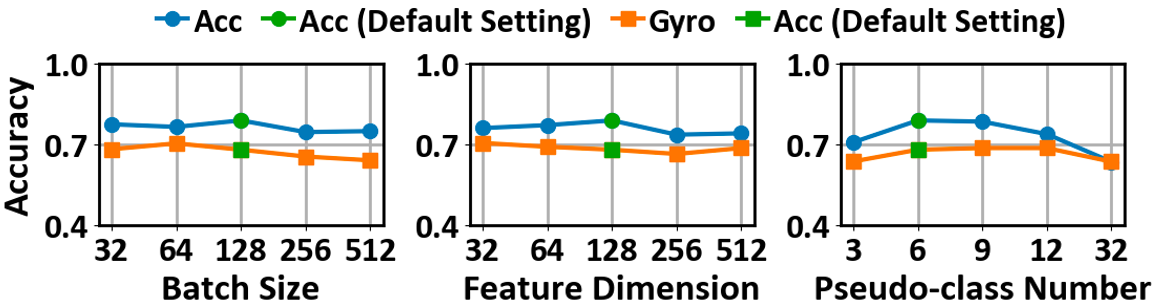}
    \caption{The impact of different parameter settings on the performance of MESEN. 
    }
    \label{fig:micro_benchmark}
\end{figure}

%% file: 9_Conclusion.tex
\section{Conclusion}
\label{sec:conclusion}
This paper proposes MESEN, a universal framework utilizing increasingly available unlabeled multimodal data to enhance unimodal HAR with few labels.
MESEN achieves effective unimodal feature extraction during the multimodal-aided pre-training stage by featuring a multi-task mechanism.
The proposed mechanism combines cross-modal feature contrastive learning and multimodal pseudo-classification aligning to exploit the correlations and relationships within multimodal data.
With the extracted effective unimodal features, MESEN then can adapt to downstream unimodal HAR with only a few labeled samples.
Our evaluation demonstrates that MESEN can significantly improve unimodal HAR performance by exploiting multimodal data.

%% file: 10_acknowledge.tex
\begin{acks}
Thank Prof. Zhenyu Yan and Dr. Xiaomin Ouyang for granting access to the Cosmo-MHAD dataset and for their insightful discussions.
Thank the anonymous reviewers and Shepherd for the valuable comments.
This research is supported by the National Natural Science Foundation of China under Grant No. U1909207, Singapore Ministry of Education Academic Research Fund Tier 1 (RG88/22), Key Research and Development Program of Zhejiang Province under Grant No. 2021C03037, and Fundamental Research Funds for the Central Universities under Grants 226-2022-00107/226-2023-00111.
\end{acks}

%% file: main.bbl

\begin{thebibliography}{45}


\ifx \showCODEN    \undefined \def \showCODEN     #1{\unskip}     \fi
\ifx \showDOI      \undefined \def \showDOI       #1{#1}\fi
\ifx \showISBNx    \undefined \def \showISBNx     #1{\unskip}     \fi
\ifx \showISBNxiii \undefined \def \showISBNxiii  #1{\unskip}     \fi
\ifx \showISSN     \undefined \def \showISSN      #1{\unskip}     \fi
\ifx \showLCCN     \undefined \def \showLCCN      #1{\unskip}     \fi
\ifx \shownote     \undefined \def \shownote      #1{#1}          \fi
\ifx \showarticletitle \undefined \def \showarticletitle #1{#1}   \fi
\ifx \showURL      \undefined \def \showURL       {\relax}        \fi
\providecommand\bibfield[2]{#2}
\providecommand\bibinfo[2]{#2}
\providecommand\natexlab[1]{#1}
\providecommand\showeprint[2][]{arXiv:#2}

\bibitem[An et~al\mbox{.}(2022)]%
        {an2022mri}
\bibfield{author}{\bibinfo{person}{Sizhe An}, \bibinfo{person}{Yin Li}, {and} \bibinfo{person}{Umit Ogras}.} \bibinfo{year}{2022}\natexlab{}.
\newblock \showarticletitle{mri: Multi-modal 3d human pose estimation dataset using mmwave, rgb-d, and inertial sensors}.
\newblock \bibinfo{journal}{\emph{Advances in Neural Information Processing Systems}}  \bibinfo{volume}{35} (\bibinfo{year}{2022}), \bibinfo{pages}{27414--27426}.
\newblock


\bibitem[Arigbabu(2020)]%
        {arigbabu2020entropy}
\bibfield{author}{\bibinfo{person}{Olasimbo~Ayodeji Arigbabu}.} \bibinfo{year}{2020}\natexlab{}.
\newblock \showarticletitle{Entropy decision fusion for smartphone sensor based human activity recognition}.
\newblock \bibinfo{journal}{\emph{arXiv preprint arXiv:2006.00367}} (\bibinfo{year}{2020}).
\newblock


\bibitem[Bi et~al\mbox{.}(2017)]%
        {bi2017familylog}
\bibfield{author}{\bibinfo{person}{Chongguang Bi}, \bibinfo{person}{Guoliang Xing}, \bibinfo{person}{Tian Hao}, \bibinfo{person}{Jina Huh}, \bibinfo{person}{Wei Peng}, {and} \bibinfo{person}{Mengyan Ma}.} \bibinfo{year}{2017}\natexlab{}.
\newblock \showarticletitle{Familylog: A mobile system for monitoring family mealtime activities}. In \bibinfo{booktitle}{\emph{2017 ieee international conference on pervasive computing and communications (percom)}}. IEEE, \bibinfo{pages}{21--30}.
\newblock


\bibitem[Chen et~al\mbox{.}(2015)]%
        {chen2015utd}
\bibfield{author}{\bibinfo{person}{Chen Chen}, \bibinfo{person}{Roozbeh Jafari}, {and} \bibinfo{person}{Nasser Kehtarnavaz}.} \bibinfo{year}{2015}\natexlab{}.
\newblock \showarticletitle{UTD-MHAD: A multimodal dataset for human action recognition utilizing a depth camera and a wearable inertial sensor}. In \bibinfo{booktitle}{\emph{2015 IEEE International conference on image processing (ICIP)}}. IEEE, \bibinfo{pages}{168--172}.
\newblock


\bibitem[Chen et~al\mbox{.}(2023)]%
        {chen2023hmgan}
\bibfield{author}{\bibinfo{person}{Ling Chen}, \bibinfo{person}{Rong Hu}, \bibinfo{person}{Menghan Wu}, {and} \bibinfo{person}{Xin Zhou}.} \bibinfo{year}{2023}\natexlab{}.
\newblock \showarticletitle{HMGAN: A Hierarchical Multi-Modal Generative Adversarial Network Model for Wearable Human Activity Recognition}.
\newblock \bibinfo{journal}{\emph{Proceedings of the ACM on Interactive, Mobile, Wearable and Ubiquitous Technologies}} \bibinfo{volume}{7}, \bibinfo{number}{3} (\bibinfo{year}{2023}), \bibinfo{pages}{1--27}.
\newblock


\bibitem[Chen et~al\mbox{.}(2020)]%
        {chen2020simple}
\bibfield{author}{\bibinfo{person}{Ting Chen}, \bibinfo{person}{Simon Kornblith}, \bibinfo{person}{Mohammad Norouzi}, {and} \bibinfo{person}{Geoffrey Hinton}.} \bibinfo{year}{2020}\natexlab{}.
\newblock \showarticletitle{A simple framework for contrastive learning of visual representations}. In \bibinfo{booktitle}{\emph{International conference on machine learning}}. PMLR, \bibinfo{pages}{1597--1607}.
\newblock


\bibitem[Chen et~al\mbox{.}(2013)]%
        {chen2013unobtrusive}
\bibfield{author}{\bibinfo{person}{Zhenyu Chen}, \bibinfo{person}{Mu Lin}, \bibinfo{person}{Fanglin Chen}, \bibinfo{person}{Nicholas~D Lane}, \bibinfo{person}{Giuseppe Cardone}, \bibinfo{person}{Rui Wang}, \bibinfo{person}{Tianxing Li}, \bibinfo{person}{Yiqiang Chen}, \bibinfo{person}{Tanzeem Choudhury}, {and} \bibinfo{person}{Andrew~T Campbell}.} \bibinfo{year}{2013}\natexlab{}.
\newblock \showarticletitle{Unobtrusive sleep monitoring using smartphones}. In \bibinfo{booktitle}{\emph{2013 7th International Conference on Pervasive Computing Technologies for Healthcare and Workshops}}. IEEE, \bibinfo{pages}{145--152}.
\newblock


\bibitem[Cho et~al\mbox{.}(2014)]%
        {cho2014properties}
\bibfield{author}{\bibinfo{person}{Kyunghyun Cho}, \bibinfo{person}{Bart Van~Merri{\"e}nboer}, \bibinfo{person}{Dzmitry Bahdanau}, {and} \bibinfo{person}{Yoshua Bengio}.} \bibinfo{year}{2014}\natexlab{}.
\newblock \showarticletitle{On the properties of neural machine translation: Encoder-decoder approaches}.
\newblock \bibinfo{journal}{\emph{arXiv preprint arXiv:1409.1259}} (\bibinfo{year}{2014}).
\newblock


\bibitem[Dawar and Kehtarnavaz(2018)]%
        {dawar2018convolutional}
\bibfield{author}{\bibinfo{person}{Neha Dawar} {and} \bibinfo{person}{Nasser Kehtarnavaz}.} \bibinfo{year}{2018}\natexlab{}.
\newblock \showarticletitle{A convolutional neural network-based sensor fusion system for monitoring transition movements in healthcare applications}. In \bibinfo{booktitle}{\emph{2018 IEEE 14th International Conference on Control and Automation (ICCA)}}. IEEE, \bibinfo{pages}{482--485}.
\newblock


\bibitem[Deldari et~al\mbox{.}(2022)]%
        {deldari2022cocoa}
\bibfield{author}{\bibinfo{person}{Shohreh Deldari}, \bibinfo{person}{Hao Xue}, \bibinfo{person}{Aaqib Saeed}, \bibinfo{person}{Daniel~V Smith}, {and} \bibinfo{person}{Flora~D Salim}.} \bibinfo{year}{2022}\natexlab{}.
\newblock \showarticletitle{COCOA: Cross Modality Contrastive Learning for Sensor Data}.
\newblock \bibinfo{journal}{\emph{Proceedings of the ACM on Interactive, Mobile, Wearable and Ubiquitous Technologies}} \bibinfo{volume}{6}, \bibinfo{number}{3} (\bibinfo{year}{2022}), \bibinfo{pages}{1--28}.
\newblock


\bibitem[Fang et~al\mbox{.}(2020)]%
        {fang2020cert}
\bibfield{author}{\bibinfo{person}{Hongchao Fang}, \bibinfo{person}{Sicheng Wang}, \bibinfo{person}{Meng Zhou}, \bibinfo{person}{Jiayuan Ding}, {and} \bibinfo{person}{Pengtao Xie}.} \bibinfo{year}{2020}\natexlab{}.
\newblock \showarticletitle{Cert: Contrastive self-supervised learning for language understanding}.
\newblock \bibinfo{journal}{\emph{arXiv preprint arXiv:2005.12766}} (\bibinfo{year}{2020}).
\newblock


\bibitem[Gatica-Perez et~al\mbox{.}(2019)]%
        {gatica2019discovering}
\bibfield{author}{\bibinfo{person}{Daniel Gatica-Perez}, \bibinfo{person}{Joan-Isaac Biel}, \bibinfo{person}{David Labbe}, {and} \bibinfo{person}{Nathalie Martin}.} \bibinfo{year}{2019}\natexlab{}.
\newblock \showarticletitle{Discovering eating routines in context with a smartphone app}. In \bibinfo{booktitle}{\emph{Adjunct Proceedings of the 2019 ACM International Joint Conference on Pervasive and Ubiquitous Computing and Proceedings of the 2019 ACM International Symposium on Wearable Computers}}. \bibinfo{pages}{422--429}.
\newblock


\bibitem[Girdhar et~al\mbox{.}(2023)]%
        {girdhar2023imagebind}
\bibfield{author}{\bibinfo{person}{Rohit Girdhar}, \bibinfo{person}{Alaaeldin El-Nouby}, \bibinfo{person}{Zhuang Liu}, \bibinfo{person}{Mannat Singh}, \bibinfo{person}{Kalyan~Vasudev Alwala}, \bibinfo{person}{Armand Joulin}, {and} \bibinfo{person}{Ishan Misra}.} \bibinfo{year}{2023}\natexlab{}.
\newblock \showarticletitle{Imagebind: One embedding space to bind them all}. In \bibinfo{booktitle}{\emph{Proceedings of the IEEE/CVF Conference on Computer Vision and Pattern Recognition}}. \bibinfo{pages}{15180--15190}.
\newblock


\bibitem[Haresamudram et~al\mbox{.}(2021)]%
        {haresamudram2021contrastive}
\bibfield{author}{\bibinfo{person}{Harish Haresamudram}, \bibinfo{person}{Irfan Essa}, {and} \bibinfo{person}{Thomas Pl{\"o}tz}.} \bibinfo{year}{2021}\natexlab{}.
\newblock \showarticletitle{Contrastive predictive coding for human activity recognition}.
\newblock \bibinfo{journal}{\emph{Proceedings of the ACM on Interactive, Mobile, Wearable and Ubiquitous Technologies}} \bibinfo{volume}{5}, \bibinfo{number}{2} (\bibinfo{year}{2021}), \bibinfo{pages}{1--26}.
\newblock


\bibitem[He et~al\mbox{.}(2022)]%
        {he2022collaborative}
\bibfield{author}{\bibinfo{person}{Shibo He}, \bibinfo{person}{Kun Shi}, \bibinfo{person}{Chen Liu}, \bibinfo{person}{Bicheng Guo}, \bibinfo{person}{Jiming Chen}, {and} \bibinfo{person}{Zhiguo Shi}.} \bibinfo{year}{2022}\natexlab{}.
\newblock \showarticletitle{Collaborative sensing in Internet of Things: A comprehensive survey}.
\newblock \bibinfo{journal}{\emph{IEEE Communications Surveys \& Tutorials}} (\bibinfo{year}{2022}).
\newblock


\bibitem[Hochreiter and Schmidhuber(1997)]%
        {hochreiter1997long}
\bibfield{author}{\bibinfo{person}{Sepp Hochreiter} {and} \bibinfo{person}{J{\"u}rgen Schmidhuber}.} \bibinfo{year}{1997}\natexlab{}.
\newblock \showarticletitle{Long short-term memory}.
\newblock \bibinfo{journal}{\emph{Neural computation}} \bibinfo{volume}{9}, \bibinfo{number}{8} (\bibinfo{year}{1997}), \bibinfo{pages}{1735--1780}.
\newblock


\bibitem[Jain and Kanhangad(2017)]%
        {jain2017human}
\bibfield{author}{\bibinfo{person}{Ankita Jain} {and} \bibinfo{person}{Vivek Kanhangad}.} \bibinfo{year}{2017}\natexlab{}.
\newblock \showarticletitle{Human activity classification in smartphones using accelerometer and gyroscope sensors}.
\newblock \bibinfo{journal}{\emph{IEEE Sensors Journal}} \bibinfo{volume}{18}, \bibinfo{number}{3} (\bibinfo{year}{2017}), \bibinfo{pages}{1169--1177}.
\newblock


\bibitem[Kim et~al\mbox{.}(2021)]%
        {kim2021wearable}
\bibfield{author}{\bibinfo{person}{Yeon-Wook Kim}, \bibinfo{person}{Kyung-Lim Joa}, \bibinfo{person}{Han-Young Jeong}, {and} \bibinfo{person}{Sangmin Lee}.} \bibinfo{year}{2021}\natexlab{}.
\newblock \showarticletitle{Wearable IMU-based human activity recognition algorithm for clinical balance assessment using 1D-CNN and GRU ensemble model}.
\newblock \bibinfo{journal}{\emph{Sensors}} \bibinfo{volume}{21}, \bibinfo{number}{22} (\bibinfo{year}{2021}), \bibinfo{pages}{7628}.
\newblock


\bibitem[Li et~al\mbox{.}(2018)]%
        {li2018co}
\bibfield{author}{\bibinfo{person}{Chao Li}, \bibinfo{person}{Qiaoyong Zhong}, \bibinfo{person}{Di Xie}, {and} \bibinfo{person}{Shiliang Pu}.} \bibinfo{year}{2018}\natexlab{}.
\newblock \showarticletitle{Co-occurrence feature learning from skeleton data for action recognition and detection with hierarchical aggregation}. In \bibinfo{booktitle}{\emph{Proceedings of the 27th International Joint Conference on Artificial Intelligence}}. \bibinfo{pages}{786--792}.
\newblock


\bibitem[Li et~al\mbox{.}(2021)]%
        {li2021contrastive}
\bibfield{author}{\bibinfo{person}{Yunfan Li}, \bibinfo{person}{Peng Hu}, \bibinfo{person}{Zitao Liu}, \bibinfo{person}{Dezhong Peng}, \bibinfo{person}{Joey~Tianyi Zhou}, {and} \bibinfo{person}{Xi Peng}.} \bibinfo{year}{2021}\natexlab{}.
\newblock \showarticletitle{Contrastive clustering}. In \bibinfo{booktitle}{\emph{Proceedings of the AAAI Conference on Artificial Intelligence}}, Vol.~\bibinfo{volume}{35}. \bibinfo{pages}{8547--8555}.
\newblock


\bibitem[Lu et~al\mbox{.}(2020)]%
        {lu2020milliego}
\bibfield{author}{\bibinfo{person}{Chris~Xiaoxuan Lu}, \bibinfo{person}{Muhamad Risqi~U Saputra}, \bibinfo{person}{Peijun Zhao}, \bibinfo{person}{Yasin Almalioglu}, \bibinfo{person}{Pedro~PB De~Gusmao}, \bibinfo{person}{Changhao Chen}, \bibinfo{person}{Ke Sun}, \bibinfo{person}{Niki Trigoni}, {and} \bibinfo{person}{Andrew Markham}.} \bibinfo{year}{2020}\natexlab{}.
\newblock \showarticletitle{milliEgo: single-chip mmWave radar aided egomotion estimation via deep sensor fusion}. In \bibinfo{booktitle}{\emph{Proceedings of the 18th Conference on Embedded Networked Sensor Systems}}. \bibinfo{pages}{109--122}.
\newblock


\bibitem[Ma et~al\mbox{.}(2019)]%
        {ma2019attnsense}
\bibfield{author}{\bibinfo{person}{Haojie Ma}, \bibinfo{person}{Wenzhong Li}, \bibinfo{person}{Xiao Zhang}, \bibinfo{person}{Songcheng Gao}, {and} \bibinfo{person}{Sanglu Lu}.} \bibinfo{year}{2019}\natexlab{}.
\newblock \showarticletitle{AttnSense: Multi-level attention mechanism for multimodal human activity recognition.}. In \bibinfo{booktitle}{\emph{IJCAI}}. \bibinfo{pages}{3109--3115}.
\newblock


\bibitem[Ma et~al\mbox{.}(2021)]%
        {ma2021unsupervised}
\bibfield{author}{\bibinfo{person}{Haojie Ma}, \bibinfo{person}{Zhijie Zhang}, \bibinfo{person}{Wenzhong Li}, {and} \bibinfo{person}{Sanglu Lu}.} \bibinfo{year}{2021}\natexlab{}.
\newblock \showarticletitle{Unsupervised human activity representation learning with multi-task deep clustering}.
\newblock \bibinfo{journal}{\emph{Proceedings of the ACM on Interactive, Mobile, Wearable and Ubiquitous Technologies}} \bibinfo{volume}{5}, \bibinfo{number}{1} (\bibinfo{year}{2021}), \bibinfo{pages}{1--25}.
\newblock


\bibitem[Malekzadeh et~al\mbox{.}(2019)]%
        {malekzadeh2019mobile}
\bibfield{author}{\bibinfo{person}{Mohammad Malekzadeh}, \bibinfo{person}{Richard~G Clegg}, \bibinfo{person}{Andrea Cavallaro}, {and} \bibinfo{person}{Hamed Haddadi}.} \bibinfo{year}{2019}\natexlab{}.
\newblock \showarticletitle{Mobile sensor data anonymization}. In \bibinfo{booktitle}{\emph{Proceedings of the international conference on internet of things design and implementation}}. \bibinfo{pages}{49--58}.
\newblock


\bibitem[{NVIDIA Corporation}({[n.\,d.]})]%
        {nano}
\bibfield{author}{\bibinfo{person}{{NVIDIA Corporation}}.} \bibinfo{year}{[n.\,d.]}\natexlab{}.
\newblock \bibinfo{title}{{Jetson Nano Developer Kit}}.
\newblock
\newblock
\urldef\tempurl%
\url{https://developer.nvidia.com/embedded/jetson-nano-developer-kit}
\showURL{%
\tempurl}


\bibitem[Ouyang et~al\mbox{.}(2022)]%
        {ouyang2022cosmo}
\bibfield{author}{\bibinfo{person}{Xiaomin Ouyang}, \bibinfo{person}{Xian Shuai}, \bibinfo{person}{Jiayu Zhou}, \bibinfo{person}{Ivy~Wang Shi}, \bibinfo{person}{Zhiyuan Xie}, \bibinfo{person}{Guoliang Xing}, {and} \bibinfo{person}{Jianwei Huang}.} \bibinfo{year}{2022}\natexlab{}.
\newblock \showarticletitle{Cosmo: contrastive fusion learning with small data for multimodal human activity recognition}. In \bibinfo{booktitle}{\emph{Proceedings of the 28th Annual International Conference on Mobile Computing And Networking}}. \bibinfo{pages}{324--337}.
\newblock


\bibitem[Paszke et~al\mbox{.}(2019)]%
        {paszke2019pytorch}
\bibfield{author}{\bibinfo{person}{Adam Paszke}, \bibinfo{person}{Sam Gross}, \bibinfo{person}{Francisco Massa}, \bibinfo{person}{Adam Lerer}, \bibinfo{person}{James Bradbury}, \bibinfo{person}{Gregory Chanan}, \bibinfo{person}{Trevor Killeen}, \bibinfo{person}{Zeming Lin}, \bibinfo{person}{Natalia Gimelshein}, \bibinfo{person}{Luca Antiga}, {et~al\mbox{.}}} \bibinfo{year}{2019}\natexlab{}.
\newblock \showarticletitle{Pytorch: An imperative style, high-performance deep learning library}.
\newblock \bibinfo{journal}{\emph{Advances in neural information processing systems}}  \bibinfo{volume}{32} (\bibinfo{year}{2019}).
\newblock


\bibitem[Rabbi et~al\mbox{.}(2018)]%
        {rabbi2018virtual}
\bibfield{author}{\bibinfo{person}{Fazlay Rabbi}, \bibinfo{person}{Taiwoo Park}, \bibinfo{person}{Biyi Fang}, \bibinfo{person}{Mi Zhang}, {and} \bibinfo{person}{Youngki Lee}.} \bibinfo{year}{2018}\natexlab{}.
\newblock \showarticletitle{When virtual reality meets internet of things in the gym: Enabling immersive interactive machine exercises}.
\newblock \bibinfo{journal}{\emph{Proceedings of the ACM on interactive, mobile, wearable and ubiquitous technologies}} \bibinfo{volume}{2}, \bibinfo{number}{2} (\bibinfo{year}{2018}), \bibinfo{pages}{1--21}.
\newblock


\bibitem[Radhakrishnan et~al\mbox{.}(2020)]%
        {radhakrishnan2020erica}
\bibfield{author}{\bibinfo{person}{Meera Radhakrishnan}, \bibinfo{person}{Darshana Rathnayake}, \bibinfo{person}{Ong~Koon Han}, \bibinfo{person}{Inseok Hwang}, {and} \bibinfo{person}{Archan Misra}.} \bibinfo{year}{2020}\natexlab{}.
\newblock \showarticletitle{ERICA: enabling real-time mistake detection \& corrective feedback for free-weights exercises}. In \bibinfo{booktitle}{\emph{Proceedings of the 18th Conference on Embedded Networked Sensor Systems}}. \bibinfo{pages}{558--571}.
\newblock


\bibitem[Reyes-Ortiz et~al\mbox{.}(2016)]%
        {reyes2016transition}
\bibfield{author}{\bibinfo{person}{Jorge-L Reyes-Ortiz}, \bibinfo{person}{Luca Oneto}, \bibinfo{person}{Albert Sam{\`a}}, \bibinfo{person}{Xavier Parra}, {and} \bibinfo{person}{Davide Anguita}.} \bibinfo{year}{2016}\natexlab{}.
\newblock \showarticletitle{Transition-aware human activity recognition using smartphones}.
\newblock \bibinfo{journal}{\emph{Neurocomputing}}  \bibinfo{volume}{171} (\bibinfo{year}{2016}), \bibinfo{pages}{754--767}.
\newblock


\bibitem[Saeed et~al\mbox{.}(2019)]%
        {saeed2019multi}
\bibfield{author}{\bibinfo{person}{Aaqib Saeed}, \bibinfo{person}{Tanir Ozcelebi}, {and} \bibinfo{person}{Johan Lukkien}.} \bibinfo{year}{2019}\natexlab{}.
\newblock \showarticletitle{Multi-task self-supervised learning for human activity detection}.
\newblock \bibinfo{journal}{\emph{Proceedings of the ACM on Interactive, Mobile, Wearable and Ubiquitous Technologies}} \bibinfo{volume}{3}, \bibinfo{number}{2} (\bibinfo{year}{2019}), \bibinfo{pages}{1--30}.
\newblock


\bibitem[Selvaraju et~al\mbox{.}(2017)]%
        {selvaraju2017grad}
\bibfield{author}{\bibinfo{person}{Ramprasaath~R Selvaraju}, \bibinfo{person}{Michael Cogswell}, \bibinfo{person}{Abhishek Das}, \bibinfo{person}{Ramakrishna Vedantam}, \bibinfo{person}{Devi Parikh}, {and} \bibinfo{person}{Dhruv Batra}.} \bibinfo{year}{2017}\natexlab{}.
\newblock \showarticletitle{Grad-cam: Visual explanations from deep networks via gradient-based localization}. In \bibinfo{booktitle}{\emph{Proceedings of the IEEE international conference on computer vision}}. \bibinfo{pages}{618--626}.
\newblock


\bibitem[Sheng et~al\mbox{.}(2022)]%
        {sheng2022facilitating}
\bibfield{author}{\bibinfo{person}{Zhiyao Sheng}, \bibinfo{person}{Huatao Xu}, \bibinfo{person}{Qian Zhang}, {and} \bibinfo{person}{Dong Wang}.} \bibinfo{year}{2022}\natexlab{}.
\newblock \showarticletitle{Facilitating Radar-Based Gesture Recognition With Self-Supervised Learning}. In \bibinfo{booktitle}{\emph{2022 19th Annual IEEE International Conference on Sensing, Communication, and Networking (SECON)}}. IEEE, \bibinfo{pages}{154--162}.
\newblock


\bibitem[Shoaib et~al\mbox{.}(2014)]%
        {shoaib2014fusion}
\bibfield{author}{\bibinfo{person}{Muhammad Shoaib}, \bibinfo{person}{Stephan Bosch}, \bibinfo{person}{Ozlem~Durmaz Incel}, \bibinfo{person}{Hans Scholten}, {and} \bibinfo{person}{Paul~JM Havinga}.} \bibinfo{year}{2014}\natexlab{}.
\newblock \showarticletitle{Fusion of smartphone motion sensors for physical activity recognition}.
\newblock \bibinfo{journal}{\emph{Sensors}} \bibinfo{volume}{14}, \bibinfo{number}{6} (\bibinfo{year}{2014}), \bibinfo{pages}{10146--10176}.
\newblock


\bibitem[Stisen et~al\mbox{.}(2015)]%
        {stisen2015smart}
\bibfield{author}{\bibinfo{person}{Allan Stisen}, \bibinfo{person}{Henrik Blunck}, \bibinfo{person}{Sourav Bhattacharya}, \bibinfo{person}{Thor~Siiger Prentow}, \bibinfo{person}{Mikkel~Baun Kj{\ae}rgaard}, \bibinfo{person}{Anind Dey}, \bibinfo{person}{Tobias Sonne}, {and} \bibinfo{person}{Mads~M{\o}ller Jensen}.} \bibinfo{year}{2015}\natexlab{}.
\newblock \showarticletitle{Smart devices are different: Assessing and mitigatingmobile sensing heterogeneities for activity recognition}. In \bibinfo{booktitle}{\emph{Proceedings of the 13th ACM conference on embedded networked sensor systems}}. \bibinfo{pages}{127--140}.
\newblock


\bibitem[Tang et~al\mbox{.}(2020)]%
        {tang2020rethinking}
\bibfield{author}{\bibinfo{person}{Wensi Tang}, \bibinfo{person}{Guodong Long}, \bibinfo{person}{Lu Liu}, \bibinfo{person}{Tianyi Zhou}, \bibinfo{person}{Jing Jiang}, {and} \bibinfo{person}{Michael Blumenstein}.} \bibinfo{year}{2020}\natexlab{}.
\newblock \showarticletitle{Rethinking 1d-cnn for time series classification: A stronger baseline}.
\newblock \bibinfo{journal}{\emph{arXiv preprint arXiv:2002.10061}} (\bibinfo{year}{2020}), \bibinfo{pages}{1--7}.
\newblock


\bibitem[Tian et~al\mbox{.}(2020)]%
        {tian2020contrastive}
\bibfield{author}{\bibinfo{person}{Yonglong Tian}, \bibinfo{person}{Dilip Krishnan}, {and} \bibinfo{person}{Phillip Isola}.} \bibinfo{year}{2020}\natexlab{}.
\newblock \showarticletitle{Contrastive multiview coding}. In \bibinfo{booktitle}{\emph{Computer Vision--ECCV 2020: 16th European Conference, Glasgow, UK, August 23--28, 2020, Proceedings, Part XI 16}}. Springer, \bibinfo{pages}{776--794}.
\newblock


\bibitem[Van~der Maaten and Hinton(2008)]%
        {van2008visualizing}
\bibfield{author}{\bibinfo{person}{Laurens Van~der Maaten} {and} \bibinfo{person}{Geoffrey Hinton}.} \bibinfo{year}{2008}\natexlab{}.
\newblock \showarticletitle{Visualizing data using t-SNE.}
\newblock \bibinfo{journal}{\emph{Journal of machine learning research}} \bibinfo{volume}{9}, \bibinfo{number}{11} (\bibinfo{year}{2008}).
\newblock


\bibitem[Vaswani et~al\mbox{.}(2017)]%
        {vaswani2017attention}
\bibfield{author}{\bibinfo{person}{Ashish Vaswani}, \bibinfo{person}{Noam Shazeer}, \bibinfo{person}{Niki Parmar}, \bibinfo{person}{Jakob Uszkoreit}, \bibinfo{person}{Llion Jones}, \bibinfo{person}{Aidan~N Gomez}, \bibinfo{person}{{\L}ukasz Kaiser}, {and} \bibinfo{person}{Illia Polosukhin}.} \bibinfo{year}{2017}\natexlab{}.
\newblock \showarticletitle{Attention is all you need}.
\newblock \bibinfo{journal}{\emph{Advances in neural information processing systems}}  \bibinfo{volume}{30} (\bibinfo{year}{2017}).
\newblock


\bibitem[Wang et~al\mbox{.}(2021)]%
        {wang2021cline}
\bibfield{author}{\bibinfo{person}{Dong Wang}, \bibinfo{person}{Ning Ding}, \bibinfo{person}{Piji Li}, {and} \bibinfo{person}{Haitao Zheng}.} \bibinfo{year}{2021}\natexlab{}.
\newblock \showarticletitle{CLINE: Contrastive Learning with Semantic Negative Examples for Natural Language Understanding}. In \bibinfo{booktitle}{\emph{Proceedings of the 59th Annual Meeting of the Association for Computational Linguistics and the 11th International Joint Conference on Natural Language Processing (Volume 1: Long Papers)}}. \bibinfo{pages}{2332--2342}.
\newblock


\bibitem[Xu et~al\mbox{.}(2023)]%
        {xu2023practically}
\bibfield{author}{\bibinfo{person}{Huatao Xu}, \bibinfo{person}{Pengfei Zhou}, \bibinfo{person}{Rui Tan}, {and} \bibinfo{person}{Mo Li}.} \bibinfo{year}{2023}\natexlab{}.
\newblock \showarticletitle{Practically Adopting Human Activity Recognition}. In \bibinfo{booktitle}{\emph{Proceedings of the 29th Annual International Conference on Mobile Computing and Networking}}. \bibinfo{pages}{1--15}.
\newblock


\bibitem[Xu et~al\mbox{.}(2021)]%
        {xu2021limu}
\bibfield{author}{\bibinfo{person}{Huatao Xu}, \bibinfo{person}{Pengfei Zhou}, \bibinfo{person}{Rui Tan}, \bibinfo{person}{Mo Li}, {and} \bibinfo{person}{Guobin Shen}.} \bibinfo{year}{2021}\natexlab{}.
\newblock \showarticletitle{Limu-bert: Unleashing the potential of unlabeled data for imu sensing applications}. In \bibinfo{booktitle}{\emph{Proceedings of the 19th ACM Conference on Embedded Networked Sensor Systems}}. \bibinfo{pages}{220--233}.
\newblock


\bibitem[Yao et~al\mbox{.}(2017)]%
        {yao2017deepsense}
\bibfield{author}{\bibinfo{person}{Shuochao Yao}, \bibinfo{person}{Shaohan Hu}, \bibinfo{person}{Yiran Zhao}, \bibinfo{person}{Aston Zhang}, {and} \bibinfo{person}{Tarek Abdelzaher}.} \bibinfo{year}{2017}\natexlab{}.
\newblock \showarticletitle{Deepsense: A unified deep learning framework for time-series mobile sensing data processing}. In \bibinfo{booktitle}{\emph{Proceedings of the 26th international conference on world wide web}}. \bibinfo{pages}{351--360}.
\newblock


\bibitem[Yao et~al\mbox{.}(2018)]%
        {yao2018sensegan}
\bibfield{author}{\bibinfo{person}{Shuochao Yao}, \bibinfo{person}{Yiran Zhao}, \bibinfo{person}{Huajie Shao}, \bibinfo{person}{Chao Zhang}, \bibinfo{person}{Aston Zhang}, \bibinfo{person}{Shaohan Hu}, \bibinfo{person}{Dongxin Liu}, \bibinfo{person}{Shengzhong Liu}, \bibinfo{person}{Lu Su}, {and} \bibinfo{person}{Tarek Abdelzaher}.} \bibinfo{year}{2018}\natexlab{}.
\newblock \showarticletitle{Sensegan: Enabling deep learning for internet of things with a semi-supervised framework}.
\newblock \bibinfo{journal}{\emph{Proceedings of the ACM on interactive, mobile, wearable and ubiquitous technologies}} \bibinfo{volume}{2}, \bibinfo{number}{3} (\bibinfo{year}{2018}), \bibinfo{pages}{1--21}.
\newblock


\bibitem[Zhang and Sawchuk(2012)]%
        {zhang2012usc}
\bibfield{author}{\bibinfo{person}{Mi Zhang} {and} \bibinfo{person}{Alexander~A Sawchuk}.} \bibinfo{year}{2012}\natexlab{}.
\newblock \showarticletitle{USC-HAD: A daily activity dataset for ubiquitous activity recognition using wearable sensors}. In \bibinfo{booktitle}{\emph{Proceedings of the 2012 ACM conference on ubiquitous computing}}. \bibinfo{pages}{1036--1043}.
\newblock


\end{thebibliography}
